\documentclass{article}
\usepackage{graphicx}
\usepackage[ruled,vlined]{algorithm2e}
\SetKwInput{KwIn}{Require}
\SetKwInput{KwOut}{Ensure}
\SetAlgoNlRelativeSize{-1}
\SetAlgoSkip{medskip}
\DontPrintSemicolon
\usepackage{listings}
\usepackage{tcolorbox}
\tcbuselibrary{breakable}
\usepackage{tabularx}
\usepackage{bbm}
\usepackage{multirow}
\usepackage{amsmath}
\usepackage{mathtools}

\usepackage{amssymb}
\usepackage[preprint]{neurips_2026}
\usepackage[utf8]{inputenc} % allow utf-8 input
\usepackage[T1]{fontenc}    % use 8-bit T1 fonts
\usepackage{hyperref}       % hyperlinks
\usepackage{url}            % simple URL typesetting
\usepackage{booktabs}       % professional-quality tables
\usepackage{amsfonts}       % blackboard math symbols
\usepackage{nicefrac}       % compact symbols for 1/2, etc.
\usepackage{microtype}      % microtypography
\usepackage{xcolor}         % colors
\definecolor{mycitecolor}{HTML}{6da7c9}
\definecolor{mblue}{HTML}{7db6ce}
\definecolor{mpurple}{HTML}{ac90d0}
\usepackage{amsmath}
\usepackage{amssymb}
\usepackage{float}
\usepackage{wrapfig}

\usepackage{algcompatible}
\usepackage{algpseudocode}
\usepackage{makecell}
\usepackage{subcaption}

\lstdefinestyle{pythonstyle}{
  language=Python,
  basicstyle=\ttfamily\small,
  keywordstyle=\color{blue},
  stringstyle=\color{green!50!black},
  commentstyle=\color{gray},
  showstringspaces=false,
  breaklines=true,
  frame=single,
  backgroundcolor=\color{gray!2}
}

\lstdefinestyle{jsonstyle}{
  basicstyle=\ttfamily\small,
  breaklines=true,
  frame=none,
  backgroundcolor=\color{gray!2},
  keywordstyle=\color{blue}\bfseries,
  stringstyle=\color{black},
  showstringspaces=false,
}

\title{COOP$^2$: Defining, Observing, and Repairing Cooperation in LLM Multi-Agent Systems}

\hypersetup{
  colorlinks=true,     % use colored text instead of boxes
  linkcolor=black,     % section links
  citecolor=mycitecolor, % citation links
  urlcolor=black       % urls
}

\author{%
  Hanqing Yang \\
  CMU
  \And
  Narjes Nourzad\thanks{
    Equal contribution.
    Contact: \texttt{hanqing3@andrew.cmu.edu}.
    CMU: Carnegie Mellon University;
    USC: University of Southern California;
    SUTD: Singapore University of Technology and Design;
    UArizona: University of Arizona.
  } \\
  USC
  \And
  Shiyu Chen\footnotemark[1] \\
  CMU
  \AND
  Marie Siew \\
  SUTD
  \And
  Jingdi Chen \\
  UArizona
  \And
  Carlee Joe-Wong \\
  CMU
}
\begin{document}

\maketitle

\begin{abstract}

Many complex tasks require extended effort, diverse capabilities, or coordinated actions beyond what a single agent can provide. 
However, simply adding more agents does not guarantee better performance, as effective cooperation depends on \textit{how} agents interact with each other and with task structure to satisfy evolving constraints over time. 
This challenge is amplified for LLM-based multi-agent systems (LLM-MAS): plans, messages, and revisions occur in natural language, whereas task progress depends on grounded environment actions. 
Current evaluations mostly treat cooperation as an \textit{implicit} ingredient of final task success, leaving both cooperation and the effect of multi-agent interaction on task dynamics difficult to study. 
We introduce COOP$^2$, an evaluation framework that grounds high-level agent \textbf{coop}eration dynamics in LLM-MAS within task progress in the environment. COOP$^2$ then defines \textbf{coop}erative tasks with verifiable cooperative requirements, allowing us to analyze how cooperation unfolds over time with respect to task progress, as well as \textit{where} and \textit{why} cooperation breaks down.
Building on this framework, we develop COOP$^2$-Repair, which predicts constraint failures from group plans and opens targeted repair channels for guided revisions.
Across two environments and three communication structures, COOP$^2$-Repair improves task success and constraint satisfaction while exposing the additional decision overhead and communication load required for repair. The project web page can be found at: \url{https://happyeureka.github.io/coop2}.
\end{abstract}

\section{Introduction}
\label{sec:intro}

% introducing LLM-MAS: why llm mas? for complex tasks
Recent progress in large language models (LLMs) has enabled agents that reason, plan, communicate, and act in interactive environments~\citep{sumers2023cognitive,durante2024agent}. 
As these agents are applied to more complex tasks that require long-horizon execution, diverse capabilities, or joint actions, many settings exceed what a \textit{single} agent can reliably accomplish.
%\carlee{give an example of how multiple agents help task completion for a long-horizon example here. The example should also illustrate that agents and the task can evolve over time (as otherwise we do not need agent re-planning)}
% For example, if agents control robots carrying boxes (e.g., in a factory setting or in a game of moving apartments \carlee{cite}), small objects can be carried independently, while large ones may require multiple agents to lift together; progress may also depend on clearing blocked pathways before other work can continue and adapting to agents' changing capabilities as they become tired. Agents must then allocate work, synchronize joint actions, and adapt to changing conditions over time.
Consider, for example, agents collaboratively building a web application from a user specification, where some components can be implemented independently, while others require cooperation across frontend, backend, database, and deployment agents. Progress may stall on failing workflows that block downstream modules, or on newly introduced bugs, dependency conflicts, and mismatches with user expectations~\citep{tran2026vibecodebenchevaluating,jimenez2024swebench,yang2024swebenchmultimodal}.  Similar dynamics arise in embodied and game-playing settings, where agents must allocate work, synchronize interdependent changes, and adapt to evolving constraints over time~\citep{nourzad2025dr,savva2019habitat}.
% Agents must then allocate work, synchronize interdependent changes, and adapt to evolving constraints over time.
% This motivates multi-agent systems in which agents must allocate work, synchronize joint actions, and revise plans as conditions evolve.\jc{moving apartment might not be a good example here, how about some applications that more robotics? or more LLM-focused cooperative tasks? Like robots in factory, or robot requiring jointly controlled joints? Or our problem solving tasks, or the agent team example?}%agents must adapt their plans over time depending on their evolving capabilities and task requirements.

%A common way to build LLM-MAS is to prescribe cooperation through workflows or hand-crafted rewards; such designs can be effective in structured settings, but they become limiting when tasks require scalable and adaptive behavior. In workflow-driven LLM-MAS, cooperation is often prescribed through predefined roles, communication protocols, or task decompositions, making agent interaction largely a consequence of system design~\citep{}. In multi-agent optimization, cooperation is often encouraged indirectly through hand-crafted rewards.
%In both cases, cooperation is engineered rather than examined. As task conditions evolve, agents must adapt, but such adaptation is difficult to fully prescribe through workflows or induce through hand-crafted rewards. 
%\noindent\textsc{The limits of prescribed cooperation.} 
LLM-MAS systems often prescribe cooperation through workflows or encourage it through hand-crafted rewards~\citep{liu2026llm, wu2024autogen}. 
While effective when task structure and expected agent behaviors are well understood, such designs become limiting for complex tasks with adaptive structure, larger teams, or cross-task generalization, since adaptive behavior under evolving conditions is difficult to fully prescribe or induce through predefined components. 
%\carlee{This is true, but I think we should move these points to the first paragraph of the paper and argue that engineered cooperation is insufficient for complex tasks (otherwise people may immediately question why we are studying agent interactions if they can be pre-specified or encouraged implicitly).} 
% In the moving-apartment task, for example, an initial plan cannot anticipate every blocked hallway, heavy object, or tired agent, and a fixed reward cannot easily specify the appropriate response in each case. These assumptions become increasingly limiting as LLM-MAS move toward more autonomous settings, sub-agent systems where agents need to define roles, allocate work, and monitor progress, and open-ended agent teams where agents must identify shared objectives and coordinate under evolving task requirements.
%\jc
% In end-to-end web application development, an initial workflow cannot anticipate every implementation failure, broken pipeline, changing interface, or hidden cross-module dependency, and a fixed reward cannot easily specify how agents should respond when these decisions become coupled~\citep{tran2026vibecodebenchevaluating,jimenez2024swebench,yang2024swebenchmultimodal}. 
These limitations grow more pronounced as LLM-MAS move toward autonomous settings where agents must define roles, allocate work, monitor progress, and adapt to evolving task requirements. 
%Studying cooperation, therefore, requires understanding how agent interactions unfold and how they shape task progress. \carlee{I would replace this last sentence with something pointing to cooperation repair: e.g., ``By making cooperation formally observable, we hypothesize that one can detect failures in agent cooperation and prompt agents to correct them, ultimately increasing task success rates.''} 
% These settings raise the promise that larger agent teams could solve increasingly complex tasks, but realizing this promise depends on whether additional agents can be effectively integrated into the evolving task process.
% Yet, simply adding more agents does not guarantee better performance \citep{kim2025towards}; %\carlee{cite the Google paper on scaling here}
Task outcomes then depend on \textit{how} agents coordinate their actions with respect to task structure, shared state, and evolving task constraints~\citep{kim2025towards}, that is, on their \textit{cooperative behavior}~\citep{chen2025five,dafoe2020open}. 
Cooperation must therefore be studied as a \textit{dynamic process} by which agents adapt their interactions to satisfy evolving task requirements, rather than as a system design.

\noindent\textsc{The cognitive--primitive structure of LLM-MAS.} Studying LLM-MAS cooperation as a process is complicated by a two-layer structure absent from classical multi-agent systems~\citep{hong2023metagpt,chen2023agentverse,zhuge2024gptswarm}.
LLMs equip agents with cognitive capabilities that operate over a \textit{high-level symbolic space}, while task progress unfolds through \textit{grounded actions in a lower-level environment}. This separation introduces asynchrony between reasoning and execution. 
Agents may spend varying amounts of time planning and communicating, while committed plans execute as a sequence of primitive actions with variable durations and uncertain outcomes. 
%During execution, incoming messages or changes in task structure may interrupt or invalidate active plans, forcing agents to return to reasoning and revise their plans at different environment steps.
%\carlee{This paragraph seems to imply that all agents need to plan. I think we need to explain why this is still a problem even if there is a central coordinator, e.g., something like: ``Even if agents are controlled by a central coordinator, they may have autonomy in deciding how to execute their assigned tasks: e.g., the coordinator may assign the task of ``help agent 1 search this area,'' but depending on how fast the agents can cover the area or what they find, this coordinator may need to adaptively re-plan task allocation, interrupting other agents in the midst of executing on the previously determined plan.''}
Even when a central coordinator assigns subtasks, worker agents may execute them at different speeds or encounter local obstacles, forcing the coordinator to  revise task allocation and interrupt agents mid-execution. Understanding cooperation in LLM-MAS, therefore, requires connecting asynchronous agent interaction to environment-level state transitions and task progress.

\begin{figure*}
    \centering
    \includegraphics[width=1\linewidth]{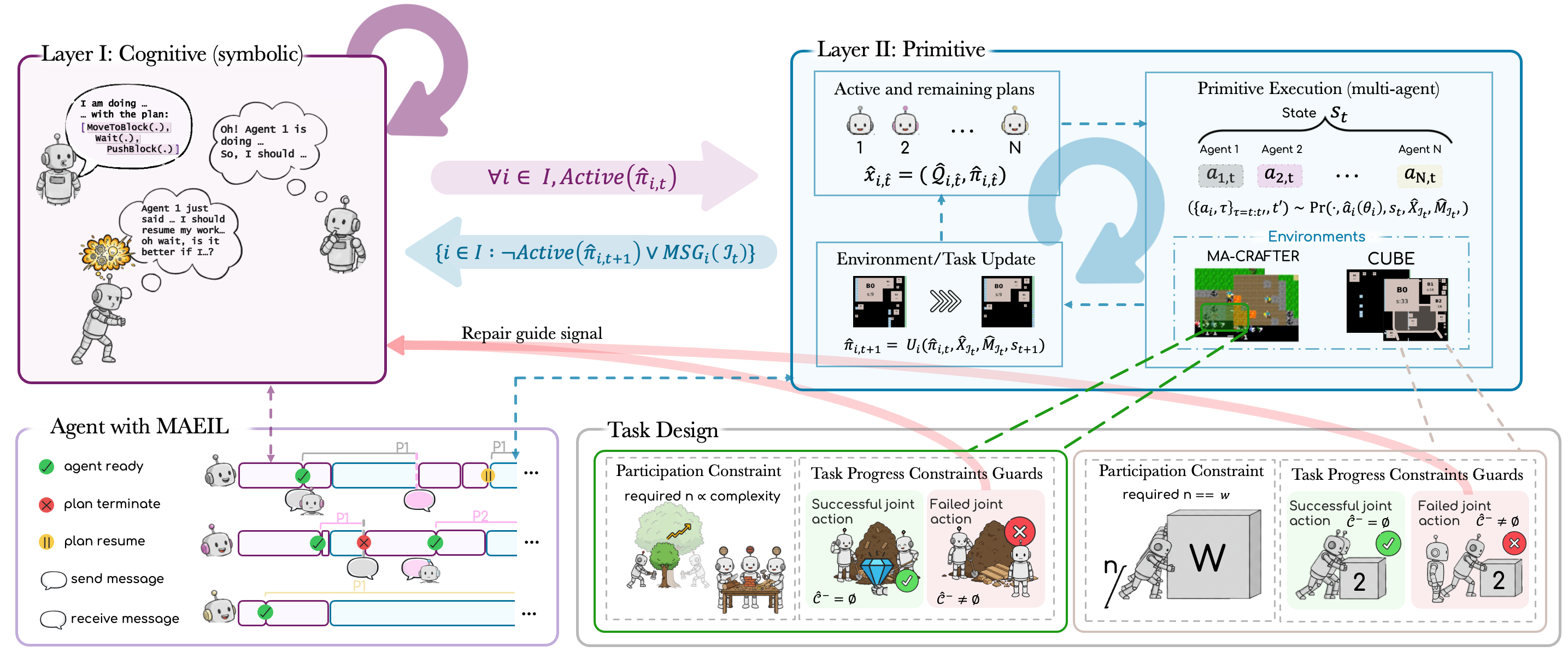}
    \caption{\textbf{Overview of COOP$^2$.} COOP$^2$ models LLM-MAS through a \textbf{cognitive--primitive interface} (top) that couples a symbolic \textcolor{mpurple}{cognitive layer}, where agents plan, communicate, interrupt, and  replan through the Multi-Agent Environment Interaction Loop (MAEIL, bottom-left), with a \textcolor{mblue}{primitive layer}, where joint actions drive environment states and task-state transitions (sec.~\ref{sec:cog_prim_interface}). \textbf{Cooperative tasks}, instantiated in MA-CRAFTER and CUBE, are guarded by participation and task-progress constraints; joint executions that satisfy all active constraints succeed while violations identify which cooperative requirements failed. These constraint signals support process-level diagnosis and feed COOP$^2$-Repair (sec.~\ref{sec:method}), which opens targeted repair channels to revise plans before primitive execution.}
    \label{fig:overall}
    \vspace{-0.2in}
\end{figure*}

% gap for evalaution
\noindent\textsc{The evaluation gap.} Even when agent interactions are connected to environment-level task progress, evaluating the cooperative process still requires formal specification of an environment's evolving cooperation requirements. Existing multi-agent evaluations, however, mostly treat cooperation as an implicit ingredient 
of final task success. 
Many LLM-MAS benchmarks evaluate outcome-based performance on reasoning-centric tasks, such as task completion, answer accuracy, or coding fixes verified by unit tests~\citep{zhu2025multiagentbench,agashe2025llm,kim2025towards,qian2024scaling}. Such outcome measures cannot reveal whether agents actually cooperated, how their interactions contributed to progress, or why task failures occurred: a system may succeed through individual effort without meaningful cooperation, or fail despite partially effective cooperation when specific task requirements go unmet. Formalizing task cooperation requirements potentially allows for these cooperation failures to be detected and repaired so as to improve task success.

%Current outcome-level evaluations thus offers limited insight into how cooperation emerges, stabilizes, or breaks down as agents interact with one another and with evolving task structure.  \carlee{I would replace this last sentence with something motivating the need for specifying cooperation constraints, e.g., ``Even if we connect agent interactions to environment-level task progress, there is therefore a need to formalize and quantify how well agents fulfill an environment's cooperation requirements, and thus how much cooperation contributes to task success.''}

% what is COOP2? defining cooperative tasks and cooperation
\noindent\textsc{Our Contributions: The COOP$^2$ framework.}  In this paper, we introduce COOP$^2$, a framework for making cooperation a first-class object of study in LLM-MAS. In doing so, we make four major contributions.  \textbf{(1) COOP$^2$ agent trace.} %By grounding agents' cognitive activities in environment state transitions, 
COOP$^2$ defines a model of agent actions that connects high-level communication and decision-making, including agents' symbolic activity, plans, messages, interruptions, and replanning; to grounded actions and task-process transitions. Thus, we can analyze the effects of cooperative dynamics on unfolding task progress. \textbf{(2) A constraint-based cooperation formalism.} COOP$^2$ then formalizes \textbf{coop}erative tasks as constraint-guarded state transitions, and \textbf{coop}eration as the process by which agents communicate, plan, revise, and act to jointly satisfy task constraints over time. We instantiate four cooperation constraints: \textit{spatial}, \textit{temporal}, \textit{participation}, and \textit{dependency}, that provide verifiable, environment-agnostic signals for diagnosing cooperative progress. \textbf{(3) COOP$^2$-Repair.} Building on this formalism, we develop COOP$^2$-Repair, a constraint-aware communication method that monitors agents' remaining plan feasibility and opens targeted repair channels by deciding when agents should communicate, who should participate, and which constraint should be addressed. \textbf{(4) Environment instantiations and validation.} Across two cooperative environments and multiple LLM backbones, team sizes, and communication structures,  COOP$^2$ supports analysis beyond final task success. COOP$^2$-Repair demonstrates how capability- and constraint-aware prediction can guide more efficient task progress, while exposing the additional decision overhead and communication load required for repair.

\section{Related Work}
\label{sec:related_work}

Recent LLM-MAS frameworks use workflows, roles, automatic topology search, or multi-agent optimization to improve collaborative task solving~\citep{gao2025survey,wu2024autogen,hong2023metagpt,zhuge2024gptswarm,chen2023agentverse}. Existing benchmarks and environments evaluate LLM agents in embodied, collaborative, or reasoning-centric settings~\citep{yang2025embodiedbench,zhu2025multiagentbench,agashe2025llm,chollet2019measure,sun-etal-2025-collab,mosquera2025can,savva2019habitat}. Classical MAS studies cooperation as an interactive process~\citep{durfee1993cooperative,yokoo2002distributed}, while single-agent LLM methods study reasoning, acting, planning, and self-correction~\citep{wei2022chain,yao2022react,wang2023plan,shinn2023reflexion}. In contrast, \textsc{COOP$^2$} standardizes cooperation itself by defining cooperative constraints, traces language-level interaction into grounded execution, measures process-level cooperation dynamics, and repairs predicted coordination failures. Due to space constraints, we provide a more comprehensive discussion of related work in Appendix~\ref{app:complete_related_work}.

\section{Cognitive--Primitive Interface for LLM-MAS}
\label{sec:cog_prim_interface}

\begin{wrapfigure}[20]{r}{0.5\textwidth}
    \centering
    \vspace{-20pt}
    \includegraphics[width=0.48\textwidth]{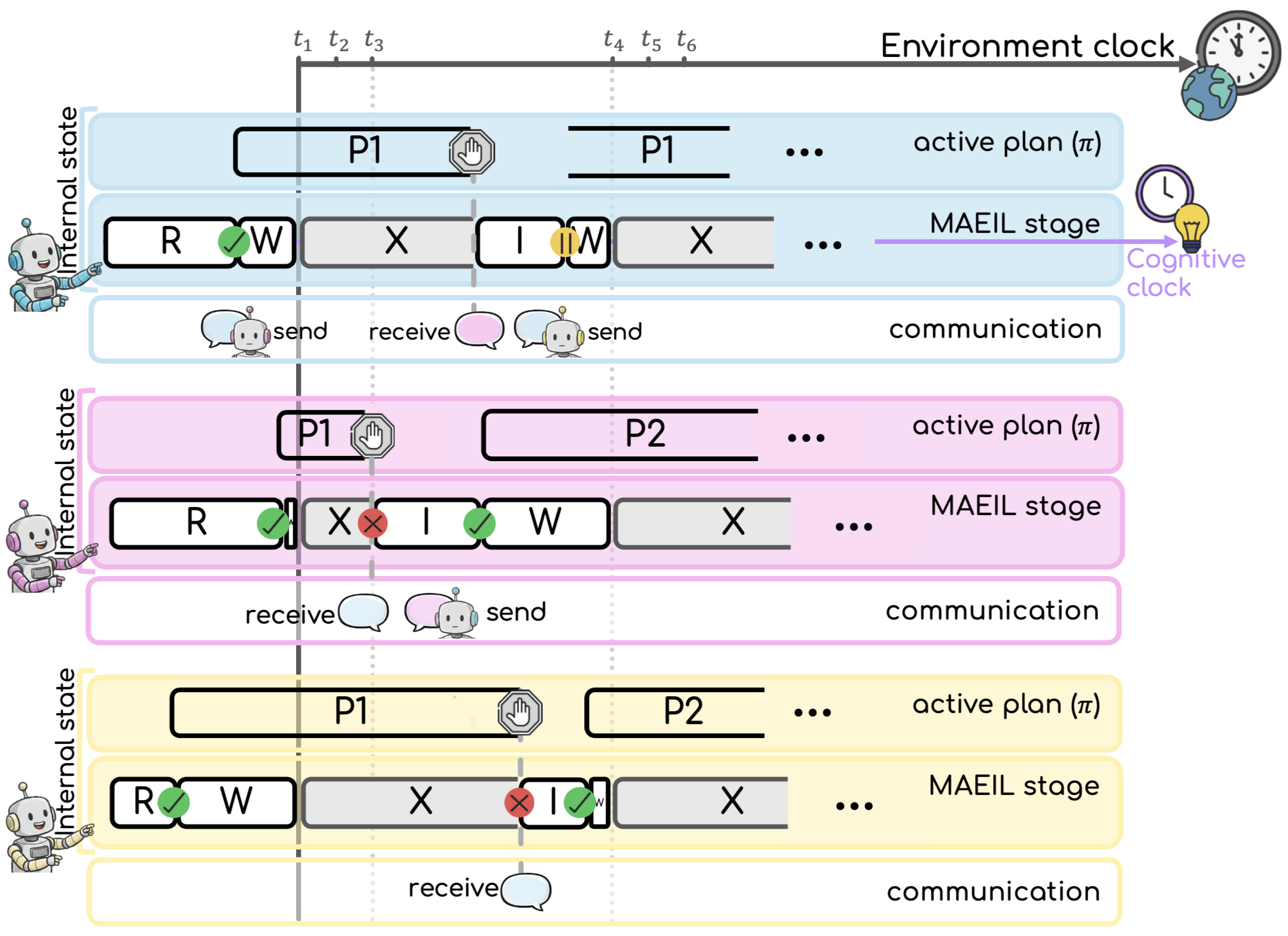}
    \caption{Decoupling cognitive and environment clocks in multi-agent execution. At each cognitive step, an agent maintains an internal state consisting of its current interaction stage and its committed symbolic plan. Plans may be newly generated (green), resumed (yellow), or terminated (red, indicating success or failure).}
    \label{fig:planing_prog}
    \vspace{-5pt}
\end{wrapfigure}
COOP$^2$ models LLM-based multi-agent systems through two coupled levels: a \textit{cognitive layer} in which agents form symbolic plans, communicate, interrupt, and replan, and a \textit{primitive layer} in which environment states evolve through primitive actions and task-state transitions. 
The interface aligns these two levels so that high-level agent interaction can be connected to primitive task progress.

\textsc{Primitive layer.}
Let $I=\{1,\dots,n\}$ denote the agent set and let $t\in\mathbb T=\{0,\dots,T\}$ index primitive environment steps. We decompose the primitive state as
% \begin{equation}
    $s_t=(\mathcal G_t,\{x_{g,t}\}_{g\in\mathcal G_t},s_t^-),$
% \end{equation}
where $\mathcal G_t$ is the active task set, $x_{g,t}$ is the state of task $g$ (including its cooperation requirements, define in Section~\ref{sec:task-design-coop}), and $s_t^-$ contains remaining environment variables that may affect execution but are not directly tracked at the task level (e.g., the presence of other objects in a navigation environment). Given a joint primitive action $a_t=(a_{1,t},\dots,a_{n,t})$, we consider its task-level effect:
%\begin{equation*}
    $\bigl(\mathcal G_{t+1},\{x_{g,t+1}\}_{g\in\mathcal G_{t+1}}\bigr)
    \xleftarrow{\,a_t\,}
    \bigl(\mathcal G_t,\{x_{g,t}\}_{g\in\mathcal G_t}\bigr)$
%\end{equation*}
which may complete active tasks, activate new tasks, or modify %states and constraints of 
existing tasks. %\carlee{do we also track completed tasks?}\hq{}

\textsc{Cognitive layer.}
The cognitive layer is the high-level space in which LLM agents reason to form and revise plans, communicate, and call tools. An agent action is represented as a tool operation $\hat a_i(\theta_i)\in\hat{\mathcal A}$, where $\hat a_i$ denotes the operation selected by agent $i$ and $\theta_i$ its arguments. 
The symbolic tool space $\hat{\mathcal A}$ may be dynamically revised, expanded, or reduced during interaction. 
To capture multi-agent dynamics in which decisions and interactions may occur asynchronously across agents, we use a cognitive index $\hat t\in\hat{\mathbb T}$,  distinct from the primitive index
$t$ and aligned to it through the grounding bridge (Eq.~\ref{eq:clock_alignment}). 
For each agent $i$, we define its cognitive activity status as
\begin{equation}
    \hat x_{i,\hat t}
    =
    (\hat{\mathcal Q}_{i,\hat t},\hat\pi_{i,\hat t}),
    \qquad
    \hat{\mathcal Q}_{i,\hat t}\in\{R,W,X,I\},
    \qquad
    \hat\pi_{i,\hat t}
    =
    \bigl(\hat\rho_{\hat\pi},
    [\hat a_{i,1}(\theta_{i,1}),\hat a_{i,2}(\theta_{i,2}),\dots]\bigr).
    \label{eq:cognitive_plan}
\end{equation}
Here, $\hat{\mathcal Q}_{i,\hat t}$ indexes the interaction stage, which is automatically updated as agents reason $(R)$ to send or receive messages and form or revise plans, wait $(W)$ for other agents to commit their plans, execute $(X)$ their plans, or are interrupted $(I)$ by other agent messages (full stage dynamics in Appendix~\ref{app:maeil}). % \carlee{we should specify which of these correspond to $R,W,X,I$}. 
The plan $\hat\pi_{i,\hat t}$ contains a task specification $\hat\rho_{\hat\pi}$, encoding the agent's current understanding of what task it is working toward, and a sequence of tool operations intended to satisfy that specification.

\textsc{Grounding bridge.}
The grounding bridge connects the cognitive and primitive layers by aligning their time scales, grounding symbolic tool operations into real-time primitive execution, and specifying when agents transition between them.

\textit{Cognitive-to-primitive clock alignment}. 
We align cognitive and primitive time by defining $\kappa:\hat{\mathbb T}\rightarrow\mathbb T$:
\begin{equation}
    \kappa(\hat t)
    =
    \max\{t\in\mathbb T:t\preceq \hat t\},
    \quad
    \mathcal I_t=[\hat t_t,\hat t_{t+1}),
    \quad
    \hat X_{\mathcal I_t}
    =
    \{\hat x_{i,\tau}\}_{i\in I,\tau\in\mathcal I_t},
    \quad
    \hat M_{\mathcal I_t}
    =
    \{m_\tau\}_{\tau\in\mathcal I_t}.
    \label{eq:clock_alignment}
\end{equation}
where $t\preceq \hat t$ means primitive step $t$ has completed by cognitive time $\hat t$, $\mathcal I_t$ is the cognitive interval between primitive steps, $\hat X_{\mathcal I_t}$ and $\hat M_{\mathcal I_t}$ collect the cognitive activity and communication events over that interval, and $m_\tau=(s_\tau,R_\tau,p_\tau)$ denotes a  communication event at cognitive time $\tau$
the sender $s_\tau$, recipient set $R_\tau$, and payload $p_\tau$. 

\textit{Cognitive-to-primitive action grounding}. 
Each tool operation is grounded by $\Gamma$ into a non-deterministic sequence of primitive actions:
\begin{equation}
    \hat a_i(\theta_i)\in\hat{\mathcal A},
    \qquad
    \bigl(\{a_{i,\tau}\}_{\tau=t:t'},t'\bigr)
    \sim
    P_{\Gamma}
    \bigl(
    \cdot
    \mid
    \hat a_i(\theta_i),
    s_t,
    \hat X_{\mathcal I_t},
    \hat M_{\mathcal I_t}
    \bigr),
    \qquad t'\ge t .
    \label{eq:action_grounding}
\end{equation}
Since the unrolling depends on the evolving primitive state, the endpoint $t'$ cannot be known ahead of time. 
As a consequence, the plan containing this tool operation may be \textit{interrupted}, \textit{revised}, \textit{resumed}, or \textit{terminated.} 

\textit{Cognitive-to-primitive interplay}. 
By the end of $\mathcal I_t$, each agent carries a step-aligned plan $\hat\pi_{i,t}$ obtained from the latest cognitive-time plan $\hat\pi_{i,\hat t}$ with $\kappa(\hat t)=t$, written as
\begin{equation}
    \hat\pi_{i,t}
    =
    \bigl(
    \hat\rho_{\hat\pi},
    [\hat a_{i,k_i}(\theta_{i,k_i}),
    \hat a_{i,k_i+1}(\theta_{i,k_i+1}),\dots]
    \bigr),
    \qquad
    \hat\pi_{i,t+1}
    =
    U_i\bigl(
    \hat\pi_{i,t},
    \hat X_{\mathcal I_t},
    \hat M_{\mathcal I_t},
    s_{t+1}
    \bigr),
    \label{eq:plan_update}
\end{equation}
where the update $U_i$ pops the selected operation if it reaches a terminal state, resumes the remaining plan if execution continues, revises the plan after interruption, or terminates it upon failure or completion. The first tool operation in the remaining list is selected for primitive execution.

Let $\mathsf C$ and $\mathsf P$ denote the cognitive and primitive layers. 
Layer transitions are given by
\begin{equation}
    \mathsf C \rightarrow \mathsf P
    \Longleftrightarrow
    \forall i\in I,\ \mathrm{Active}(\hat\pi_{i,t}),
    \quad
    \mathsf P \rightarrow \mathsf C
    \Longleftrightarrow
    \{i\in I:
    \neg\mathrm{Active}(\hat\pi_{i,t+1})
    \vee
    \mathrm{MSG}_i(\mathcal I_t)\}
    \neq\varnothing, 
    \label{eq:layer_transitions}
\end{equation}
where $\mathrm{Active}$ indicates a non-terminated plan available for execution, and $\neg\mathrm{Active}$ covers plans that complete, fail, or become infeasible under the updated primitive state. 
$\mathrm{MSG}_{i}(\mathcal I_t)$ indicates that agent $i$ receives a message during $\mathcal I_t$, so the nonempty set in the second condition identifies the agents that re-enter the cognitive layer asynchronously. The active plans of agents in $I$ are grounded through $\Gamma$ to form the joint action $a_t=(a_{1,t},\dots,a_{n,t})$ that drives the task progress.

Together, the primitive layer, cognitive layer, and grounding bridge expose how symbolic activity over $\hat{\mathbb T}$ produces task-state changes over $\mathbb T$, providing the substrate on which we define and analyze cooperation in the next section.

\section{Cooperative Tasks and Cooperation Processes}
\label{sec:task-design-coop}

The cognitive--primitive interface grounds symbolic tool operations into primitive actions that change task structure. 
Building on this interface, we define cooperation as a process over task states, agents capability, and cooperative constraint satisfaction.

\textsc{Cooperative task.}
We model a cooperative task as a multi-agent task whose progress is guarded by cooperative constraint satisfaction. 
For each active task $g\in\mathcal G_t$, let $x_{g,t}\in\mathcal X_g$ denote its task state, $\mathcal I_{g,t}\subseteq I$ the agents participating in $g$, and $Z_{g,t}=\{z_{i,t}\}_{i\in\mathcal I_{g,t}}$ their group capability,  where \(z_{i,t}\) represents the capability state of agent \(i\). We define
\begin{equation}
        p_{g,t}^{c}
    =
    \Pr\!\left(
    C_{g,t}^{c} \mid x_{g,t},Z_{g,t}
    \right),
    \qquad
    c\in\mathcal C_g=\{\Delta t,\Delta \ell,n,d\}.
    \label{eq:constraint_likelihood}
\end{equation}
Here $C_{g,t}^{c}$ denotes the task-specific cooperative constraint of type $c$ for task $g$ at step $t$, and $p_{g,t}^{c}$ is the probability that the participating group satisfies this constraint. 
The set $\mathcal C_g$ contains four cooperative constraint types: temporal, spatial or relational, participation or capability, and dependency constraints, respectively, each acting as a prerequisite for task progress.
For instance, \textit{temporal} constraints may require agents to simultaneously take the same action (e.g., to lift an object), \textit{spatial or relational} constraints may require agents to be in specific locations at the same time, \textit{participation or capability} constraints require agent capabilities needed to accomplish a task (e.g., agents with minimum combined context lengths needed to split and read a long document), and \textit{dependency} constraints may require agents to have access to certain tools to complete a task.
%\carlee{For example, \textit{temporal} constraints may represent agents needing to simultaneously take the same action (e.g., to lift an object), \textit{spatial or relational} constraints may represent agents needing to be in specific locations at the same time, \textit{participation or capability} constraints represent agent capabilities needed to accomplish a task (e.g., agents with minimum combined context lengths needed to split and read a long document), and \textit{dependency} constraints may represent agents needing access to certain tools to complete a task.}
These constraints are general for describing cooperative decision-making across environments and team size, and provide verifiable signals for evaluating cooperation in terms of how constraint satisfaction shapes task progress.

\textsc{Cooperation process.}
We define the cooperation process at two levels. At the single-task level, task progress is monitored over the task's active span under grounded multi-agent execution:
\begin{equation}
    x_{g,t+1}
    \sim
    \Pr(\cdot \mid x_{g,t}, \{p_{g,t}^{c}\}_{c\in\mathcal C_g}),
    \quad
    \phi_{g,t}
    =
    (x_{g,t}, \mathcal I_{g,t}, Z_{g,t}, \{p_{g,t}^{c}\}_{c\in\mathcal C_g}),
    \quad
    t \in T_g.
    \label{eq:task_cooperation_process}
\end{equation}
Here \(T_g=[t_g^{\mathrm{start}},t_g^{\mathrm{end}}]\) denotes the active span of task \(g\), from its generation until \(x_{g,t}\) reaches a terminal state (completion, failure, or removal). The evolution of \(\phi_{g,t}\) over \(t\in T_g\) defines the task-level cooperation process. At the episode level, we collect all active tasks and their cooperative states as 
\begin{equation}
    \mathcal G_t
    =
    \{g:t\in T_g\},
    \qquad
    \Phi_t
    =
    \left(
    \mathcal G_t,
    \{\phi_{g,t}\}_{g\in\mathcal G_t}
    \right),
    \qquad
    L_t
    =
    (\hat X_{\mathcal I_t},\hat M_{\mathcal I_t},a_t,\Phi_t,\Phi_{t+1}).
    \label{eq:episode_cooperation_process}
\end{equation}
Here \(\mathcal G_t\) is the active task set, \(\Phi_t\) is the episode-level cooperative state, and \(L_t\) logs the transition from cognitive-layer activity \((\hat X_{\mathcal I_t},\hat M_{\mathcal I_t})\) to primitive-layer action and task transitions \((a_t,\Phi_t,\Phi_{t+1})\).

\section{COOP$^2$-Repair}
\label{sec:method}

COOP$^2$-Repair uses the COOP$^2$ trace to guide communication before primitive execution. 
At each primitive step $t$, agents first operate in the cognitive layer until all active agents commit a plan.
% Agents without a committed active plan, or agents interrupted by a message or repair request, update their cognitive states through reasoning, communication, and decision making until they commit a plan. 
% Once all active agents are ready, 
COOP$^2$-Repair then groups committed plans by their task specifications and predicts whether each group is likely to satisfy the cooperative constraints required by its intended task. 

% If any predicted constraint satisfaction is low for a task that is not on cooldown, a temporary repair channel is opened for that task, and affected agents communicate and may replan.
% After the task is repaired, it is placed on cooldown, during which COOP$^2$-Repair does not open another repair channel for the same task. \carlee{this paragraph can be integrated into the discussion below for space}

% \begin{figure}[H]
%     \centering
%     \includegraphics[width=0.9\linewidth]{repair-flow.png}
%     \caption{COOP$^2$-Repair across cognitive and primitive layers.}
%     \label{fig:coop2_repair_flow}
% \end{figure}

\textsc{Feasibility check.}
Following Eq.~\eqref{eq:plan_update}, each step-aligned plan $\hat\pi_{i,t}$ contains a task specification and a remaining sequence of tool operations. 
In COOP$^2$-Repair, we take the task specification to be the intended task index, so $\hat\rho_{\hat\pi_{i,t}}\in\mathcal G_t$. 
The group pursuing task $g$ and the corresponding group plans are
\begin{equation}
    \mathcal I_{g,t}
    =
    \{i\in I:\hat\rho_{\hat\pi_{i,t}}=g\},
    \qquad
    \hat\Pi_{g,t}
    =
    \{\hat\pi_{i,t}\}_{i\in\mathcal I_{g,t}} .
    \label{eq:task_group_plans}
\end{equation}
COOP$^2$-Repair predicts the constraint likelihoods in Eq.~\eqref{eq:constraint_likelihood} under these plans, denoted by $\widehat p_{g,t}^{c}$ for $c\in\mathcal C_g$. 
The predicted failing constraints are
% \begin{equation}
$    \widehat{\mathcal C}_{g,t}^{-}
    =
    \{c\in\mathcal C_g:\widehat p_{g,t}^{c}<\tau_c\},$
    % \qquad
    with thresholds
    $\tau_c\in[0,1]$.
    % \label{eq:failing_constraints}
% \end{equation}
If $\widehat{\mathcal C}_{g,t}^{-}\neq\varnothing$, the group pursuing task $g$ is predicted to be unlikely to complete it, and $\widehat{\mathcal C}_{g,t}^{-}$ identifies which cooperative constraints are expected to fail. 
% However, predicted failure triggers repair only when task $g$ is not on cooldown.

\textsc{Adaptive repair communication.}
When failure is predicted for a task $g$ 
%that is not on cooldown, 
COOP$^2$-Repair opens a temporary repair channel initialized with the group pursuing $g$:
\begin{equation}
    \mathcal R_{g,t}
    =
    \mathcal I_{g,t},
    \qquad
    \mathrm{ctx}_{g,t}
    =
    \bigl(
    g,
    \mathcal I_{g,t},
    \widehat{\mathcal C}_{g,t}^{-}
    \bigr).
    \label{eq:repair_channel}
\end{equation}
Here $\mathrm{ctx}_{g,t}$ identifies the task, the group, and the cooperative constraints that the group's remaining plans are expected to violate. Agents in $\mathcal R_{g,t}$ communicate and replan around the identified task and failure reason, and may contact agents outside the current channel. Contacted agents are interrupted and added to $\mathcal R_{g,t}$. 
The resulting messages become part of $\hat M_{\mathcal I_t}$, and plan revisions are captured by the active-plan update $U_i$ in Eq.~\eqref{eq:plan_update}.
After repair, task $g$ is placed on \textit{cooldown}, during which COOP$^2$-Repair does not open another repair channel for the same task. 
Agents interrupted during a repair re-enter the cognitive layer to update their plans before the next primitive step.

COOP$^2$ is intended as a general framework rather than a proposal for a specific learning model; its core contribution is to show that the COOP$^2$ representation can support both \textit{diagnosis} and interaction \textit{guidance}. COOP$^2$-Repair instantiates this idea with lightweight heuristic estimators. Learning a general constraint-satisfaction predictor is a promising direction and is left for future work.

\section{Cooperative Environment Instantiations}
\label{sec:coop-env}

We instantiate COOP$^2$ in two multi-agent environments with distinct cooperative task structures.
In both environments, each task $g$ has a score $r(g)$ proportional to the number and difficulty of its cooperative constraints.
\begin{wrapfigure}{r}{0.42\textwidth}
    \centering
    \vspace{-8pt}
    \includegraphics[width=1\linewidth]{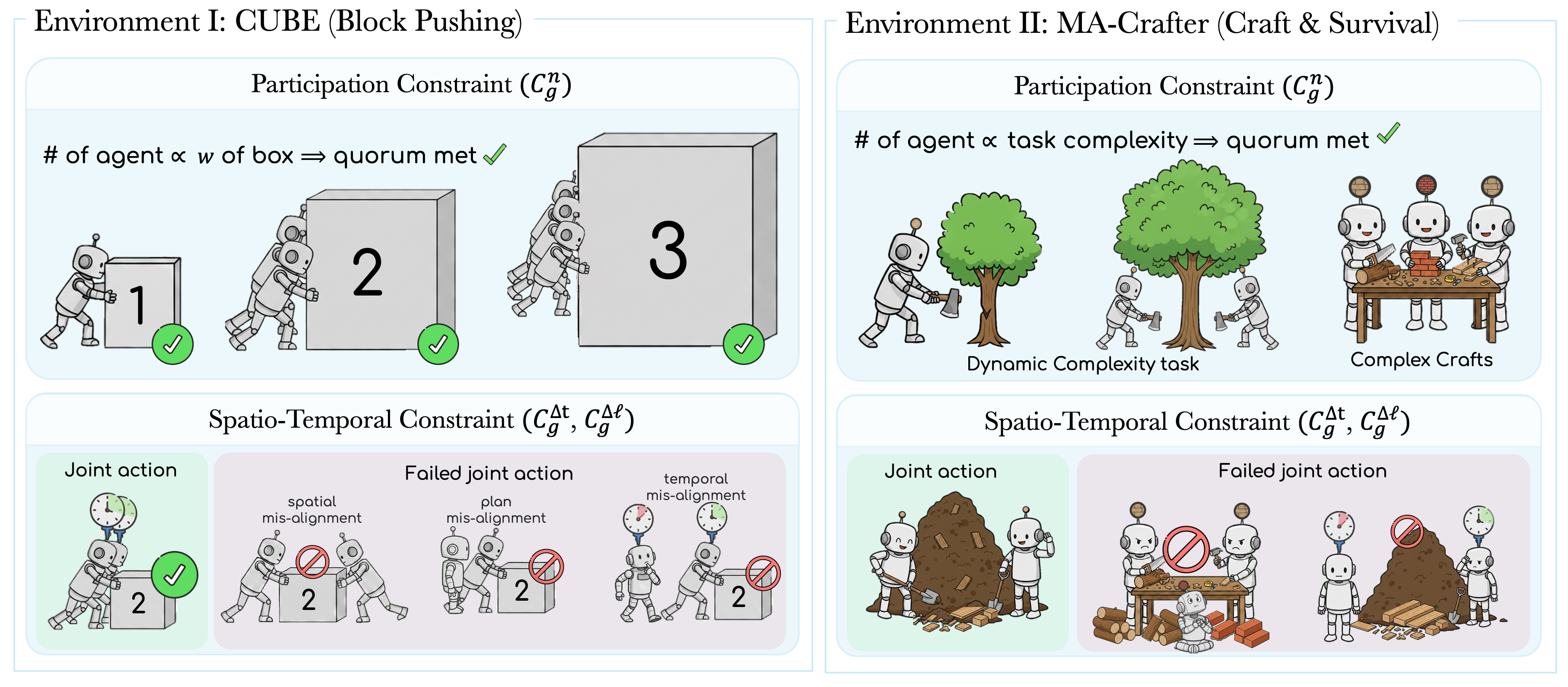}
    \caption{Two cooperative environment instantiations. \textbf{CUBE}: agents push weighted blocks where heavier blocks require more participants in spatial alignment. \textbf{MA-Crafter}: agents collect resources whose difficulty scales with prerequisites, tools, and required participants. }
    \label{fig:env_instantiations}
    \vspace{-30pt}
\end{wrapfigure}
Agents aim to maximize the total score of completed tasks within a bounded episode, which terminates when either the environment-step budget $T$ or the wall-clock budget $\hat T$ is reached, whichever comes first.
The wall-clock budget makes cognitive activity and cooperation overhead consequential: time spent reasoning, communicating, waiting, or replanning reduces the remaining opportunity for task completion.
The episode score is defined over the effective horizon induced by the environment-step and wall-clock budgets:
\begin{equation*}
       R(\tau)=
    \sum_{t=0}^{T^\star}
    \sum_{g\in\mathcal G_t}
    r(g)\,\mathrm{Done}(g),
    \quad
    T^\star=\min\{T,\kappa(\hat T)\}.
\end{equation*}

\textsc{Environment instantiations.}
We build on \textbf{MA-Crafter}~\citep{yang2025llm}, a multi-agent extension of Crafter~\citep{hafner2021benchmarking}, and \textbf{CUBE}~\citep{yangcube}, a cooperative block-pushing environment for LLM agents, shown in Figure~\ref{fig:env_instantiations}.

\begin{wraptable}{r}{0.6\textwidth}
\centering
\vspace{-15pt}
\caption{Cooperation constraint specifications for task $g$ at time $t$ in MA-Crafter and CUBE.}
\label{tab:env_constraints}
\small
\begin{tabular}{rcc}
\toprule
Constraint & \textbf{MA-Crafter} & \textbf{CUBE} \\
\midrule
$C_{\mathrm{n}}^{i}(g,t)$ 
& $\mathrm{cap}_i(t)\succeq \mathrm{cap}(g,t)$ 
& $\mathrm{cap}_i(t)$ \\
$C_{\Delta \ell}^{i}(g,t)$ 
& $\lVert \mathcal x_{i,t}-y_g\rVert \le d$ 
& $\min_{y\in \mathcal{Y}_{g}}\lVert \mathcal x_{i,t}-y\rVert \le 1$ \\
$C_{\Delta t}^{i}(g,t)$ 
& $a_{i,t}=\texttt{collect}(g)$ 
& $a_{i,t}=\texttt{push}(b(g))$ \\
\bottomrule
\end{tabular}
\vspace{-0.8em}
\end{wraptable}

In \textbf{MA-Crafter}, agents explore, collect resources, and craft tools through a technology tree.
We define each collectible resource active at time $t$ as a task $g\in\mathcal G_t$.
Advanced resources receive higher scores because they require more prerequisites, stronger tools, and more valid participants.
As summarized in Table~\ref{tab:env_constraints}, an agent counts as a valid participant only if it has the required capability, is within distance $d$ of the resource, and executes the corresponding \texttt{collect} action.
This setting emphasizes capability acquisition, dependency satisfaction, and synchronized execution. In \textbf{CUBE}, agents push weighted blocks to a goal region.
We treat each available block side at time $t$ as a task $g\in\mathcal G_t$, corresponding to pushing block $b(g)$ from side $s(g)\in\mathcal S(b(g))$, with participation threshold $p(g)=w_{b(g)}$.
Blocks with larger weights receive higher scores because they impose stronger spatial-alignment and participation constraints.
An agent counts as a valid participant only if it is adjacent to the relevant side of the block and executes \texttt{push}(b(g)).
This setting emphasizes spatial alignment, simultaneous action, and multi-agent participation.
Environment details are provided in Appendices~\ref{app:ma-crafter} and~\ref{app:cube}. Both environments use the task participation constraint $p(g)$ to determine how many agents must satisfy each required constraint. Specifically, $N(C_c)(g,t)=\sum_{i\in I}\mathbb{I}[C_c^i(g,t)]$ is the number of agents that satisfy constraint $c$ for task $g$ at step $t$. %We verify task progress from execution traces by checking whether enough agents satisfy the required capability, spatial, and temporal constraints within the effective episode horizon.
In the next section, we explain our design of experiments that use COOP$^2$ to examine agent cooperation in these environments.

\section{Experiment Design}
\label{sec:experiments}
We design experiments that exercise COOP$^2$ across multiple agent backbones, communication structures, and team sizes, allowing us to evaluate cooperation beyond final task performance and to measure the cost and quality of cooperation under different system conditions.

\textsc{Settings.}
We vary four factors: environment, LLM backbone, communication structure, and team size. We evaluate the two environments described in Sec.~\ref{sec:coop-env}. For backbones, we compare GPT-5.4-mini, GPT-5.4, and Llama-4-Scout-17B-16E, covering smaller and larger frontier models as well as an open-source model, to test how model scale and type affect cooperative progress, inference time, and failure modes. 
 For communication, we compare three settings: \textit{Individual}, where agents act independently; \textit{Centralized}, where a leader coordinates workers; and \textit{Chain}, where agents discuss in order before committing plans. 
 We evaluate both 3-agent and 6-agent teams to study how cooperation changes with team size and communication structure. This gives $36$ unique settings in total each repeated $5$ times; we report mean and standard deviation.

\textsc{Recorded outcomes.}
For each run, we record task progress, cognitive activity, communication, and constraint satisfaction from the COOP$^2$ specification defined in Sec.~\ref{sec:task-design-coop}. We summarize cooperation with eight aggregate quantities: Score (cumulative task value), Steps (episode length), and Score/Step (task value per step), which measure task progress; Plans/Agent (average committed plans per agent), which measures cognitive activity; Msg. (total messages exchanged), Intr. (interruptions), and Total Dec. (wall-clock time spent reasoning or handling interruptions), which measure communication; and finally constraint violation deficits by type. Constraint violations are computed over targeted task-attempt events and report average deficits for the constraints defined in Sec.~\ref{sec:coop-env}; lower values indicate better constraint satisfaction. Full metric definitions are provided in Appendix~\ref{app:metrics}.

\section{Main Results}
\label{sec:experiments}
We focus the main result table on \textsc{MA-Crafter}, which contains richer dependency constraints, a larger action space, and a more open-ended structure. Table~\ref{tab:mcrafter-results} reports the \textsc{MA-Crafter} results; the corresponding CUBE results are provided in Appendix~\ref{app:extra-trace}.

\begin{table*}[h]
\centering
\caption{\textsc{MA-Crafter} recorded outcomes for different agent-model settings and communication structures. Values are mean \(\pm\) standard deviation over five runs. Arrows indicate better direction when a clear direction exists; bold marks the best value within each agent-model setting.}
\label{tab:mcrafter-results}
\footnotesize
\setlength{\tabcolsep}{2.0pt}
\renewcommand{\arraystretch}{1.05}

\resizebox{\textwidth}{!}{
\begin{tabular}{lrrrrrrrrrr}
\toprule
Structure
& \makecell{Score\\$\uparrow$}
& \makecell{Steps}
& \makecell{Score/\\Step $\uparrow$}
& \makecell{Plans/\\Agent}
& \makecell{Msg.}
& \makecell{Intr.}
& \makecell{Total\\Dec.}
& \makecell{Spat.\\Viol. $\downarrow$}
& \makecell{Temp.\\Viol. $\downarrow$}
& \makecell{Dep.\\Viol. $\downarrow$} \\
\midrule

\multicolumn{11}{l}{\textbf{3-agent GPT-5.4-mini}} \\
Individual  & $\mathbf{146.8 \pm 141.6}$ & $117.4 \pm 13.0$ & $\mathbf{1.26 \pm 1.29}$ & $42.8 \pm 1.2$ & $0.0 \pm 0.0$ & $0.0 \pm 0.0$ & $196.1 \pm 8.1$ & $0.10 \pm 0.07$ & $0.01 \pm 0.02$ & $\mathbf{0.23 \pm 0.23}$ \\
Centralized & $39.4 \pm 24.3$ & $88.4 \pm 12.1$ & $0.45 \pm 0.30$ & $33.7 \pm 0.9$ & $88.4 \pm 6.9$ & $34.6 \pm 5.6$ & $251.9 \pm 5.3$ & $0.12 \pm 0.08$ & $0.01 \pm 0.01$ & $0.57 \pm 0.17$ \\
Chain      & $29.4 \pm 5.7$ & $61.8 \pm 11.9$ & $0.49 \pm 0.10$ & $33.3 \pm 1.5$ & $57.4 \pm 3.0$ & $48.0 \pm 4.9$ & $200.0 \pm 3.7$ & $\mathbf{0.05 \pm 0.05}$ & $\mathbf{0.00 \pm 0.00}$ & $0.44 \pm 0.26$ \\

\midrule
\multicolumn{11}{l}{\textbf{3-agent Llama-Scout}} \\
Individual  & $\mathbf{30.4 \pm 8.0}$ & $62.6 \pm 7.1$ & $0.48 \pm 0.11$ & $33.3 \pm 2.2$ & $0.0 \pm 0.0$ & $0.0 \pm 0.0$ & $242.9 \pm 9.6$ & $\mathbf{0.04 \pm 0.04}$ & $0.01 \pm 0.02$ & $0.24 \pm 0.19$ \\
Centralized & $24.0 \pm 6.6$ & $59.4 \pm 3.9$ & $0.41 \pm 0.11$ & $26.4 \pm 2.5$ & $50.4 \pm 10.4$ & $13.6 \pm 7.0$ & $244.1 \pm 15.5$ & $0.08 \pm 0.09$ & $0.01 \pm 0.02$ & $0.25 \pm 0.11$ \\
Chain      & $18.4 \pm 5.9$ & $29.4 \pm 8.7$ & $\mathbf{0.63 \pm 0.09}$ & $18.7 \pm 5.5$ & $29.6 \pm 7.3$ & $19.4 \pm 5.2$ & $179.6 \pm 35.7$ & $0.11 \pm 0.11$ & $\mathbf{0.00 \pm 0.00}$ & $\mathbf{0.03 \pm 0.06}$ \\

\midrule
\multicolumn{11}{l}{\textbf{3-agent GPT-5.4}} \\
Individual  & $\mathbf{809.0 \pm 405.8}$ & $99.8 \pm 10.0$ & $\mathbf{8.22 \pm 4.25}$ & $17.8 \pm 1.2$ & $0.0 \pm 0.0$ & $0.0 \pm 0.0$ & $170.5 \pm 7.1$ & $0.43 \pm 0.06$ & $0.04 \pm 0.05$ & $0.24 \pm 0.09$ \\
Centralized & $266.8 \pm 174.1$ & $68.2 \pm 3.9$ & $3.96 \pm 2.67$ & $13.5 \pm 0.3$ & $36.8 \pm 7.0$ & $19.4 \pm 2.9$ & $220.7 \pm 12.0$ & $\mathbf{0.36 \pm 0.08}$ & $\mathbf{0.00 \pm 0.01}$ & $0.11 \pm 0.07$ \\
Chain      & $181.6 \pm 84.5$ & $54.2 \pm 3.9$ & $3.37 \pm 1.58$ & $14.3 \pm 0.5$ & $26.8 \pm 1.3$ & $22.0 \pm 2.1$ & $201.3 \pm 7.3$ & $0.37 \pm 0.05$ & $0.01 \pm 0.02$ & $\mathbf{0.07 \pm 0.06}$ \\

\midrule
\multicolumn{11}{l}{\textbf{6-agent GPT-5.4-mini}} \\
Individual  & $86.8 \pm 34.5$ & $82.2 \pm 4.2$ & $1.06 \pm 0.45$ & $33.4 \pm 2.1$ & $0.0 \pm 0.0$ & $0.0 \pm 0.0$ & $322.4 \pm 8.2$ & $\mathbf{0.11 \pm 0.04}$ & $\mathbf{0.00 \pm 0.01}$ & $0.38 \pm 0.27$ \\
Centralized & $\mathbf{141.8 \pm 89.2}$ & $63.4 \pm 8.3$ & $\mathbf{2.18 \pm 1.31}$ & $23.7 \pm 0.6$ & $127.4 \pm 38.2$ & $61.4 \pm 16.7$ & $401.9 \pm 51.3$ & $0.13 \pm 0.05$ & $0.02 \pm 0.02$ & $\mathbf{0.25 \pm 0.12}$ \\
Chain      & $27.4 \pm 6.1$ & $22.6 \pm 1.3$ & $1.20 \pm 0.22$ & $16.8 \pm 1.2$ & $75.2 \pm 6.0$ & $61.2 \pm 9.9$ & $332.9 \pm 18.1$ & $0.13 \pm 0.08$ & $0.06 \pm 0.06$ & $0.38 \pm 0.13$ \\

\midrule
\multicolumn{11}{l}{\textbf{6-agent Llama-Scout}} \\
Individual  & $32.6 \pm 8.0$ & $53.2 \pm 3.3$ & $0.61 \pm 0.12$ & $35.2 \pm 3.0$ & $0.0 \pm 0.0$ & $0.0 \pm 0.0$ & $498.5 \pm 39.5$ & $\mathbf{0.09 \pm 0.04}$ & $0.02 \pm 0.04$ & $0.30 \pm 0.20$ \\
Centralized & $\mathbf{50.8 \pm 6.5}$ & $49.6 \pm 4.3$ & $1.03 \pm 0.18$ & $27.0 \pm 2.8$ & $115.2 \pm 52.8$ & $43.2 \pm 27.9$ & $484.5 \pm 33.6$ & $0.18 \pm 0.12$ & $0.01 \pm 0.01$ & $0.16 \pm 0.12$ \\
Chain      & $19.8 \pm 11.9$ & $14.2 \pm 3.8$ & $\mathbf{1.24 \pm 0.60}$ & $13.1 \pm 1.5$ & $60.6 \pm 6.2$ & $49.4 \pm 6.4$ & $386.0 \pm 12.5$ & $0.13 \pm 0.10$ & $\mathbf{0.00 \pm 0.00}$ & $\mathbf{0.05 \pm 0.09}$ \\

\midrule
\multicolumn{11}{l}{\textbf{6-agent GPT-5.4}} \\
Individual  & $\mathbf{691.4 \pm 303.5}$ & $68.4 \pm 3.7$ & $\mathbf{9.92 \pm 3.99}$ & $11.7 \pm 1.3$ & $0.0 \pm 0.0$ & $0.0 \pm 0.0$ & $215.0 \pm 9.7$ & $0.47 \pm 0.04$ & $0.01 \pm 0.01$ & $0.17 \pm 0.08$ \\
Centralized & $230.4 \pm 167.0$ & $46.0 \pm 3.0$ & $4.83 \pm 3.33$ & $9.8 \pm 0.4$ & $58.0 \pm 6.9$ & $35.8 \pm 5.1$ & $326.5 \pm 16.5$ & $0.43 \pm 0.06$ & $0.02 \pm 0.02$ & $0.06 \pm 0.07$ \\
Chain      & $34.6 \pm 14.8$ & $21.2 \pm 3.7$ & $1.63 \pm 0.63$ & $8.0 \pm 0.5$ & $36.6 \pm 2.0$ & $30.8 \pm 2.7$ & $349.9 \pm 17.0$ & $\mathbf{0.25 \pm 0.04}$ & $\mathbf{0.00 \pm 0.00}$ & $\mathbf{0.00 \pm 0.00}$ \\

\bottomrule
\end{tabular}
}
\end{table*}

\textbf{Stronger models progress better.}
GPT-5.4 achieves substantially higher Score and Score/Step than GPT-5.4-mini and Llama-Scout, both smaller models, in most matched settings. GPT-5.4 reaches a Score roughly $5\times$ that of GPT-5.4-mini ($809.0$ vs.\ $146.8$) and over $25\times$ that of Llama-Scout ($30.4$). The gap persists at 6 agents, with GPT-5.4 outperforming GPT-5.4-mini by $\sim 8\times$ and Llama-Scout by $\sim 20\times$.
This suggests that stronger backbones produce more grounded and executable plans that better account for task feasibility, satisfy capability requirements, and adapt from low-level resource collection toward higher-value tasks as capabilities evolve. 

\textbf{More agents are not always better.}
Increasing team size from 3 to 6 agents does not monotonically improve Score or Score/Step. Although larger teams provide more capacity, they also enlarge the error space through collection conflicts and decision overhead. 
Even for GPT-5.4 in Individual mode, where there is no communication by definition, scaling from 3 to 6 agents reduces Score by $\sim 15\%$ ($809.0 \to 691.4$), as additional agents compete for the same resources without coordinating. In the Centralized and Chain communication structures, the 6 agents exchange nearly twice as many messages as the 3 agents for most settings, with corresponding increases in interruptions.
Under the wall-clock budget \(\hat{\mathcal{T}}\), this overhead can reduce completed environment steps, especially when agents are interrupted during execution.

\textbf{Communication is not always cooperation.} The Centralized and Chain structures introduce more messages, interruptions, and decision time, but this additional process activity does not uniformly improve outcomes. For weaker backbones, Centralized communication can recover or improve Score over Individual baselines: 6-agent GPT-5.4-mini gains $\sim 60\%$ ($86.8 \to 141.8$). For the strongest backbone, communication overhead instead reduces Score: 3-agent GPT-5.4 drops by $\sim 67\%$ under Centralized ($809.0 \to 266.8$), as the model's independent plans are already grounded enough that synchronization adds cost without proportional benefit. Decision time confirms this overhead: 6-agent Centralized configurations require $\sim 25$--$50\%$ more Total Dec.\ than Individual.

\textbf{Constraint violations expose distinct failure modes.} Aggregate scores conceal which cooperative requirements break down. Dependency violations track model strength: GPT-5.4-mini suffers severe dependency failures (up to $93\%$ in 3-agent Centralized), while GPT-5.4 reduces these by an order of magnitude (down to $\sim 7\%$--$11\%$ in matched settings). Spatial violations, in contrast, remain substantial across all backbones, reaching $36$--$43\%$ even for GPT-5.4. This suggests spatial coordination is a failure mode largely separate from capability and dependency reasoning. Temporal violations, in contrast, stay low across the board ($<5\%$).

These aggregate outcomes help reveal cooperation by comparing models, team sizes, and communication structures, but they do not show how successes and failures arise over time. We next use COOP$^2$ traces to inspect how cooperation unfolds over time and how it correlates to task progress.

\subsection{COOP$^2$ Trace}
\begin{figure}[t]
    \centering
    \includegraphics[width=1\linewidth]{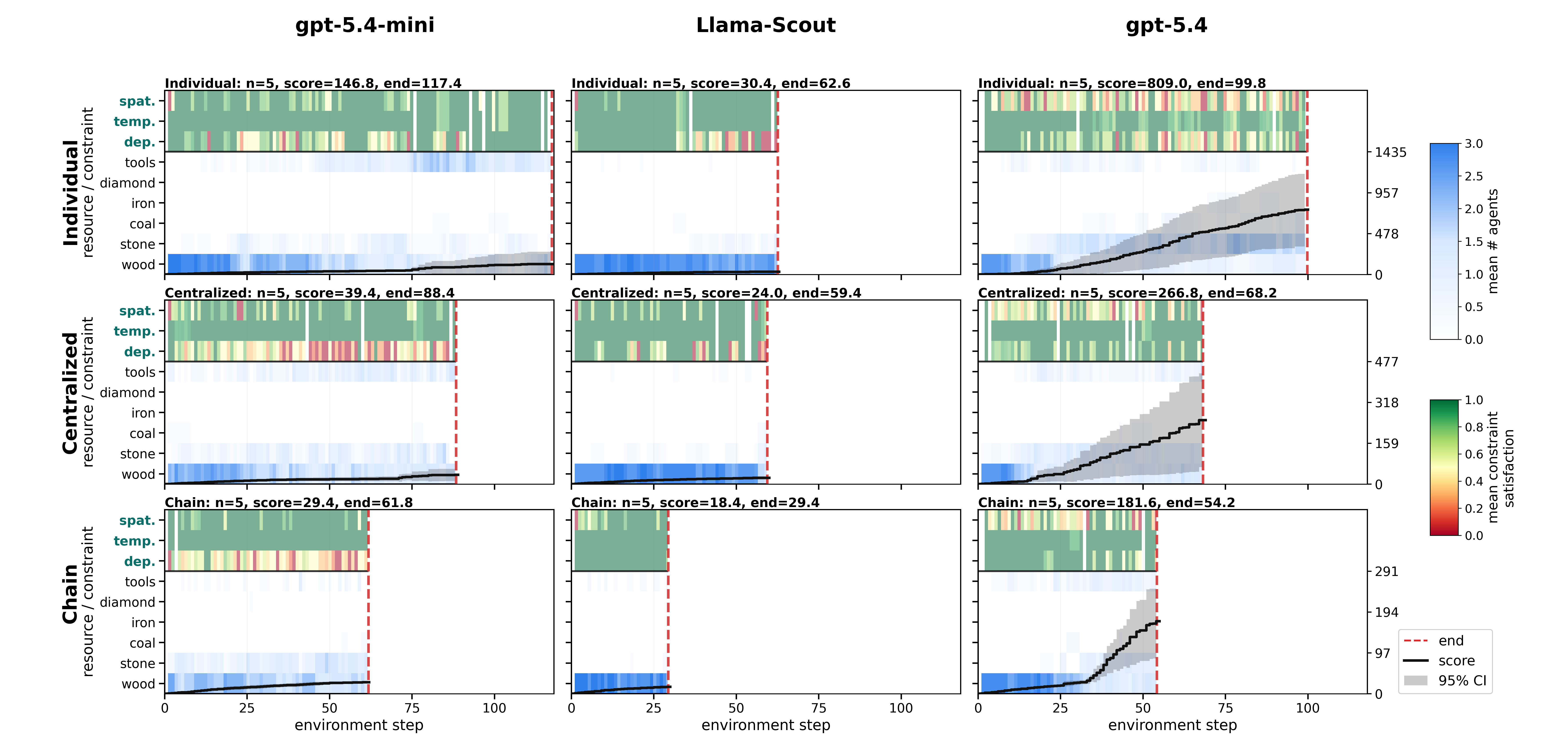}
    \caption{COOP$^2$ process traces for MA-Crafter with 3-agent teams, averaged over five runs. Columns compare backbones, and rows compare communication structures. The blue heatmap in each cell shows the number of agents whose active plans target each resource or tool at each environment step. The black curve shows cumulative team score with shaded 95\%  confidence intervals across runs, and the red dashed line marks the mean episode termination step under the cognitive-time budget $\hat{\mathcal{T}}$. The top of each cell reports observed spatial, temporal, and dependency satisfaction for attempted collection actions, indicating when and where cooperation succeeds or breaks down during execution.}
    \label{fig:process_traces_macrafter}
        \vspace{-0.3in}
\end{figure}
Figure~\ref{fig:process_traces_macrafter} illustrates an example process-level view enabled by COOP$^2$. While the trace can be further elaborated to inspect per-agent or per-task behavior, here we visualize agents' active task focus, observed constraint satisfaction at collection attempts, and cumulative team score over time. Our goal is to explain how the aggregate outcomes in Table~\ref{tab:mcrafter-results} arise from the underlying execution dynamics.

\textbf{Individual reveals model capability.}
The Individual structure does not require communication: agents avoid message passing and plan interruptions, independently forming and executing plans. This allows them to complete more environment steps within the \(\hat{T}\) budget, creating more opportunities for resource collection. However, the traces show that smaller models often spend these steps on immediately available resources, especially wood, rather than progressing through the prerequisite chain ($\text{wood} \to \text{stone} \to \text{coal} \to \text{iron} \to \text{diamond}, \text{ plus tools}$). GPT-5.4 instead shifts from low-level resources toward tools and higher-value resources, suggesting that stronger backbones produce more grounded plans and better adapt objectives as agent capabilities evolve.

\textbf{Centralized requires a capable coordinator.}
In the Centralized setting, GPT-5.4-mini shows more incomplete constraint satisfaction, especially for dependency-rich tasks: the leader may assign tasks that workers cannot yet execute, and workers may follow these assignments despite missing tools or prerequisites. GPT-5.4 instead shifts earlier toward advanced resources and shows fewer dependency failures, suggesting that centralized communication is useful only when the coordinator can identify feasible shared objectives and route them around constraint requirements.

\textbf{Chain aligns plans but delays execution.}
The Chain structure shows more similar score progression across episodes because later agents condition on earlier messages and make more aligned task choices. However, ordered discussion also increases waiting, since the environment cannot advance until the chain completes, and agents tend to follow earlier speakers, making them less likely to shift toward new resources or higher-value tasks. As a result, Chain often completes fewer environment steps, especially for slower backbones, which limits task progress under the wall-clock budget.

The traces reveal a tradeoff obscured by aggregate outcomes: \textbf{planning quality versus execution opportunity}. Stronger models produce more grounded plans, but slower reasoning can reduce completed environment steps under the cognitive-time budget. COOP$^2$ traces expose this tradeoff by linking cognitive activity, grounded execution, and constraint satisfaction over time.

\subsection{COOP$^2$-Repair}
\begin{wrapfigure}{r}{0.6\textwidth}
    \centering
    \vspace{-40pt}
    \includegraphics[width=1\linewidth]{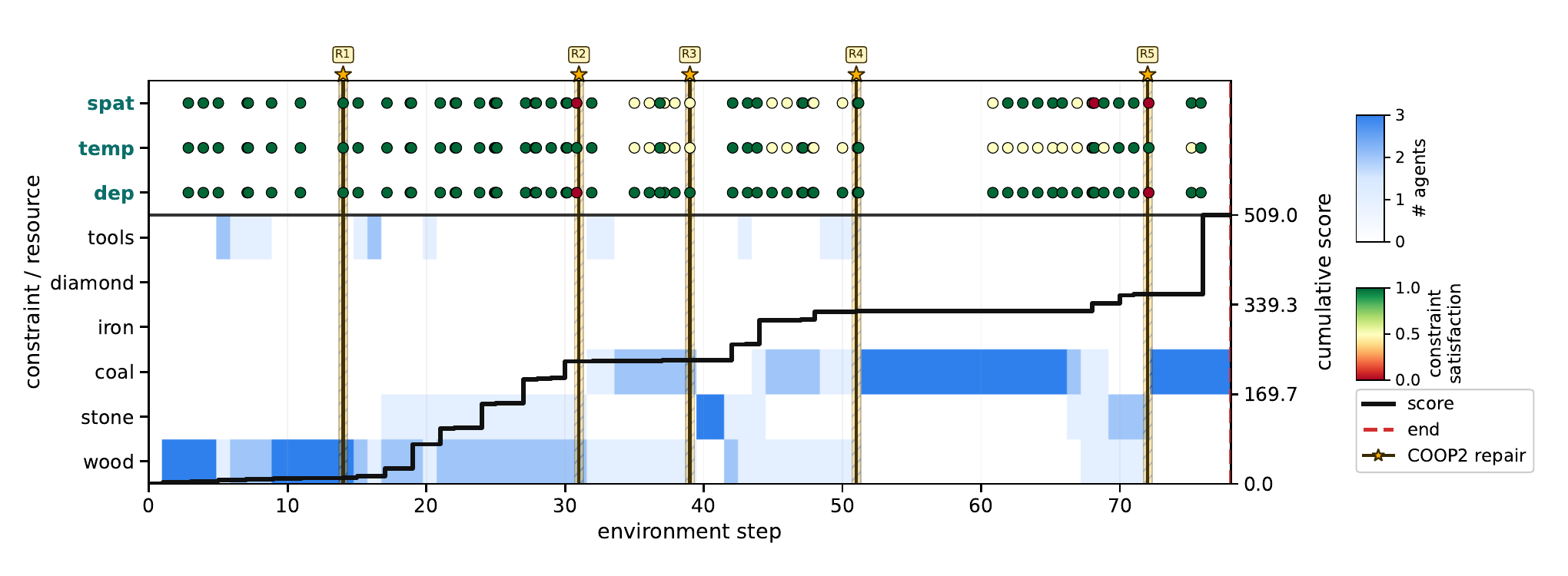}
    \caption{COOP$^2$-Repair process trace for a centralized 3-agent GPT-5.4-mini run. Gold markers indicate steps where COOP$^2$ predicts an upcoming constraint failure and opens a targeted repair round. The top rows show observed satisfaction across the spatial, temporal, and dependency constraints.}
    \label{fig:repair_case_study}
    \vspace{-10pt}
\end{wrapfigure}

Figure~\ref{fig:repair_case_study} shows a centralized 3-agent GPT-5.4-mini run with \textsc{COOP$^2$-Repair}. COOP$^2$-Repair predicts and monitors whether agents' joint plans are likely to satisfy the constraints of their target tasks given current capabilities, and opens targeted repair rounds when failure is predicted. The first repair event triggers when agents have accumulated enough resources to advance but their plans remain focused on low-level collection; the subsequent shift in active plans from wood collection toward tool crafting reflects this intervention. After this round, agents begin collecting stone, and later repair guidance helps them unlock more advanced resources such as coal, producing the sharp increase in cumulative score near the end of the episode.

\section{Conclusion}\label{sec:conclusion}
We introduce $\mathrm{COOP}^2$, an evaluation framework for studying cooperation in LLM-based multi-agent systems by making it an observable process. By exposing the interplay of cooperation signals in agents' \textbf{cognitive activity}, \textbf{environment execution}, and \textbf{cooperative constraints}, $\mathrm{COOP}^2$ shifts evaluation beyond end-task success toward interpretable cooperation dynamics and customizable metrics over interaction traces, surfacing tradeoffs between planning quality and execution opportunity that aggregate metrics obscure. Building on this representation, $\mathrm{COOP}^2$-Repair predicts constraint failures and opens targeted repair channels, enabling improved cooperative task solving. %While we propose a concrete set of cooperation metrics, there is no single canonical metric for cooperation quality. 
COOP$^2$ makes cooperation itself an object of study in LLM-MAS through grounded, process-level interaction dynamics and task design.

\begin{ack}
% Use unnumbered first-level headings for the acknowledgments. All acknowledgments

% Do {\bf not} include this section in the anonymized submission, only in the final paper. You can use the \texttt{ack} environment provided in the style file to automatically hide this section in the anonymized submission.
This work was supported in part by the Office of Naval Research under grant N000142412073 and the National Science Foundation under grants CNS-2533813 and CNS-2312761. Marie Siew was supported by the SUTD-MOE Early Career Award under the Singapore Ministry of Education START Scheme.
In addition, we would like to thank Tania Lorido-Botran for her early discussions and contributions to the project, particularly on cooperation among LLM agents.

\end{ack}

%\section*{References}

{\small
\bibliography{references, ref_cube}
\bibliographystyle{plainnat}
}
% References follow the acknowledgments in the camera-ready paper. Use unnumbered first-level heading for
% the references. Any choice of citation style is acceptable as long as you are
% consistent. It is permissible to reduce the font size to \verb+small+ (9 point)
% when listing the references.
% Note that the Reference section does not count towards the page limit.
% \medskip

% {
% \small

% [1] Alexander, J.A.\ \& Mozer, M.C.\ (1995) Template-based algorithms for
% connectionist rule extraction. In G.\ Tesauro, D.S.\ Touretzky and T.K.\ Leen
% (eds.), {\it Advances in Neural Information Processing Systems 7},
% pp.\ 609--616. Cambridge, MA: MIT Press.

% [2] Bower, J.M.\ \& Beeman, D.\ (1995) {\it The Book of GENESIS: Exploring
%   Realistic Neural Models with the GEneral NEural SImulation System.}  New York:
% TELOS/Springer--Verlag.

% [3] Hasselmo, M.E., Schnell, E.\ \& Barkai, E.\ (1995) Dynamics of learning and
% recall at excitatory recurrent synapses and cholinergic modulation in rat
% hippocampal region CA3. {\it Journal of Neuroscience} {\bf 15}(7):5249-5262.
% }

%%%%%%%%%%%%%%%%%%%%%%%%%%%%%%%%%%%%%%%%%%%%%%%%%%%%%%%%%%%%

\appendix{}
\section*{Appendix}
The appendix is organized into four clusters: \textbf{Cluster A} (Framework Details) extends the formal framework from Sections~\ref{sec:cog_prim_interface}-\ref{sec:method} with MAEIL stage dynamics, complete metric definitions, and extended related work. \textbf{Cluster B} (Implementation Details) describes the threaded execution loop and its scalability to large teams, the structured LLM interface, the role prompts for each communication structure, and timeline visualizations. \textbf{Cluster C} (Environments) documents the two cooperative environments, MA-Crafter and CUBE described in section~\ref{sec:coop-env}, including their task structures, action spaces, and environment prompts. \textbf{Cluster D} (Additional Results) provides additional COOP$^2$ traces and qualitative annotated examples of agent communication
, traces beyond those in the main paper and qualitative annotated examples of agent communication.

\section{Framework details}
This cluster extends the formal framework introduced in  Sections~\ref{sec:cog_prim_interface} - \ref{sec:method}  with detailed dynamics, full metric definitions, and a comprehensive positioning against prior work.

\subsection{Extended Related Works}
\label{app:complete_related_work}

We provide extended discussion of how COOP$^2$ relates to prior work along three dimensions: how LLM-MAS are built and structured, how cooperation has been formalized and abstracted, and how multi-agent systems are evaluated.

\noindent\textbf{LLM-MAS frameworks and multi-agent environments.} Recent work has explored LLM-MAS through human-defined workflows, automatic workflow discovery, and multi-agent optimization~\citep{gao2025survey}. LLM-MAS frameworks such as AutoGen~\citep{wu2024autogen}, MetaGPT~\citep{hong2023metagpt}, GPT-Swarm~\citep{zhuge2024gptswarm}, and AgentVerse~\citep{chen2023agentverse} prescribe cooperation implicitly through workflows, roles, or prompts. Multi-agent environments such as Collab-Overcooked~\citep{sun-etal-2025-collab}, Melting Pot~\citep{mosquera2025can}, and Habitat~\citep{savva2019habitat} define cooperative success through environment-specific rules and dynamics. While these systems support a wide range of cooperative tasks, the specification of \textit{when} cooperation is required and \textit{how} it is satisfied varies across systems and environments, making it difficult to compare cooperative behavior across settings or to disentangle cooperation from individual task performance. COOP$^2$ complements these systems by introducing a constraint-guarded representation of cooperative tasks that makes cooperation requirements explicit and verifiable across environments.

\noindent\textbf{Cooperative decision-making, classical and recent.} Classical work on cooperative distributed problem solving views cooperation as a temporally extended interactive process among agents~\citep{durfee1993cooperative,yokoo2002distributed}. More recently, single-agent LLM methods such as Chain-of-Thought~\citep{wei2022chain}, ReAct~\citep{yao2022react}, Plan-and-Solve~\citep{wang2023plan}, and Reflexion~\citep{shinn2023reflexion} provide cognitive abstractions for reasoning, acting, planning, and self-correction, but operate over a single agent's deliberation rather than multi-agent cooperation. Existing LLM-MAS frameworks focus on high-level symbolic interaction~\citep{wu2024autogen,hong2023metagpt,zhuge2024gptswarm,chen2023agentverse}, while embodied environments emphasize low-level execution and state transitions~\citep{sun-etal-2025-collab,mosquera2025can,savva2019habitat}. This separation between the symbolic and grounded views of multi-agent behavior leaves a gap in connecting language-level coordination to environment-level task progress, which COOP$^2$'s cognitive--primitive interface is designed to bridge.

\noindent\textbf{Evaluation of multi-agent systems.} Several LLM-MAS evaluation benchmarks have been proposed, including EmbodiedBench~\citep{yang2025embodiedbench}, MultiAgentBench~\citep{zhu2025multiagentbench}, and related evaluation suites~\citep{agashe2025llm,chollet2019measure,kim2025towards,qian2024scaling}. These evaluations rely primarily on reasoning-centric or outcome-based tasks such as task completion, answer accuracy, or unit-test completion. While such measures are useful for comparing systems, they cannot distinguish true cooperation from individual success or localize when and why coordination fails during execution. COOP$^2$ provides a complementary view that exposes cooperative dynamics during execution, allowing outcome-based comparisons to be interpreted alongside process-level signals.

\subsection{MAEIL: Interaction Loop and Stage Transitions}\label{app:maeil}
In this section, we further explain the multi-agent environment interaction loop (MAEIL) and the events that trigger transitions between its stages. As introduced in before, at any cognitive time $\hat{t}$, each agent $i$ maintains an internal state $ x_{i,\hat{t}} = (\mathcal Q_{i,\hat{t}}, \pi_{i,\hat{t}})$,
where $\mathcal Q_{i,\hat{t}}$ denotes the current MAEIL stage of agent $i$, and $\pi_{i,\hat{t}}$ denotes its active plan. We begin with a single agent interacting with the environment, and then extend the discussion to multiple agents that reason asynchronously while acting in a shared, stepwise environment. Communication is subsequently introduced as an explicit interaction mechanism among agents.

\paragraph{Single-agent case.} When only one embodied agent is present, the agent alternates between a \textit{reasoning stage} $\mathsf{R}$ and an \textit{execution stage} $\mathsf{X}$, such that $\mathcal Q_{i,\hat{t}}\in{\{\mathsf{R},\mathsf{X}\}}$. 
During $\mathsf{R}$, the agent may reason in any style, invoke any available tools, and take an arbitrary amount of cognitive time to produce a plan $\pi_{i,\hat{t}}$. 
{The agent executes a plan in the environment and returns to reasoning once execution completes, following interaction pattern resembling ReAct \citep{yao2022react}.}

\paragraph{Multiple agents without communication.} When multiple embodied agents act in a shared environment without communication, an additional wait stage $\mathsf{W}$ is required, so that $\mathcal Q_{i,\hat{t}}\in{\{\mathsf{R},\mathsf{X},\mathsf{W}\}}$. 
Since the environment advances only when all agents maintain an active plan, an agent that has committed its plan but is waiting for others, transitions into $\mathsf{W}$. 
\[
i:\;
\{\mathsf{R},\mathsf{X}\} \;\xrightarrow{\exists i\in I_{\mathrm{active}}:\ \neg\mathrm{Ready}}\; \mathsf{W} \;\xrightarrow{\forall i\in I_{\mathrm{active}}:\ \mathrm{Ready}}\; \mathsf{X}.
\]

\paragraph{Multiple agents with communication.} When communication is enabled, agents may receive messages at any time, including during execution. 
To model this, we introduce an  interrupt stage $\mathsf{I}$, yielding $\mathcal Q_{i,\hat t}\in \{\mathsf{R},\mathsf{X},\mathsf{W},\mathsf{I}\}$. 
Upon receiving a message, an agent transitions to $\mathsf{I}$, processes the incoming information, and decides whether to resume its current plan or replan. Agents in stages $\mathsf{R}$ and $\mathsf{I}$ may send messages, while agents in any stage may receive messages:
\[
i_2:\{\mathsf{R},\mathsf{I}\}
\;\xRightarrow{\textsc{Msg}_{i_2\to i_1}(\hat{t})}\;
i_1:\{\mathsf{R},\mathsf{X},\mathsf{W},\mathsf{I}\}.
\]
{Message arrival interrupts waiting or execution and moves the agent to $\mathsf{I}$:}
\[
i_1:\;
\{\mathsf{W},\mathsf{X}\}
\;\xrightarrow{\textsc{Msg}_{i_2\to i_1}(\hat{t})}\;
\mathsf{I}
\xrightarrow{\textsc{Ready}(i_1,\hat{t}')}
\mathsf{W},
\]
where the transition is defined for the receiving agent $i$.  After processing the message, the agent enters $\mathsf{W}$ once it maintains an active plan, either by resuming the previous plan or by committing a new one. Figure~\ref{fig:planing_prog} shows an instance of multiple communicating agents and their MAEIL transitions.

\subsection{Metrics}\label{app:metrics}

At the aggregate level, we report Score, Steps, Score/Step, Plans/Agent, Msg., Intr., Total Dec., and constraint violation deficits. These quantities are computed from the COOP$^2$ trace defined in Sec.~\ref{sec:task-design-coop}, specifically the step-aligned cognitive activity $\hat{X}_{I_t}$and message trace $\hat{M}_{I_t}$ over each interval.
%, grounded joint action $a_t$, and task-process states $\Phi_t$.

Let \(N\) denote the number of agents. We aggregate planning and communication over cognitive intervals as
\begin{equation}
    \mathrm{Plans/Agent}
=
\frac{1}{N}
\sum_{i=1}^{N}
\sum_{t=1}^{T}
\mathbf{1}\{\hat{\pi}_{i,\hat{t}} \text{ is logged in } I_t\},
\qquad
\mathrm{Msg.}
=
\sum_{t=1}^{T} |\hat{M}_{I_t}|.
\end{equation}

Interruption frequency and decision time are computed from the MAEIL stage component $\hat{Q}_{i,\hat{t}}$, with \(\Delta \hat{t}_{i,\hat{t}}\) denoting the duration of agent \(i\)'s cognitive step at time \(\hat{t}\):
\begin{equation}
    \mathrm{Intr.}
=
\sum_{i=1}^{N}
\sum_{t=1}^{T}
\sum_{\hat{t}\in I_t}
\mathbf{1}
[
\hat{Q}_{i,\hat{t}}=I
],
\qquad
\mathrm{Total\ Dec.}
=
\sum_{i=1}^{N}
\sum_{t=1}^{T}
\sum_{\hat{t}\in I_t}
\mathbf{1}
[
\hat{Q}_{i,\hat{t}}\in\{R,I\}
]
\Delta \hat{t}_{i,\hat{t}}.
\end{equation}

Constraint violations are aggregated over targeted task-attempt events. An attempt event \(e_g\) occurs when one or more agents execute the action associated with task \(g\) (e.g., \texttt{collect}(g)
in MA-Crafter)
For each attempt event \(e_g\) targeting task \(g\) with participation threshold \(p(g)\), and for each constraint type \(c\), we define
\begin{equation}
    v_c(e_g)
=
1-\min\{s_c(e_g),1\},
\qquad
\mathrm{Viol}_c
=
\frac{1}{|E_c|}
\sum_{e_g\in E_c}
v_c(e_g),
\end{equation}

where \(s_c(e_g)\) is the satisfaction score for constraint \(c\) at event \(e_g\), and \(E_c\) is the set of evaluated attempt events for constraint type \(c\), as defined in Sec.~\ref{sec:coop-env}.
 %cluster 1: rel work, metric, maeil

\newpage

\section{Implementation Details}
This cluster describes the implementation behind COOP$^2$'s experimental pipeline, including  the scalability design, the asynchronous execution loop, the structured LLM interface, and the prompts used to instantiate each communication structure.

% \newpage
\subsection{Scalable Design}
\begin{figure}[H]
    \centering
    \includegraphics[width=1\linewidth]{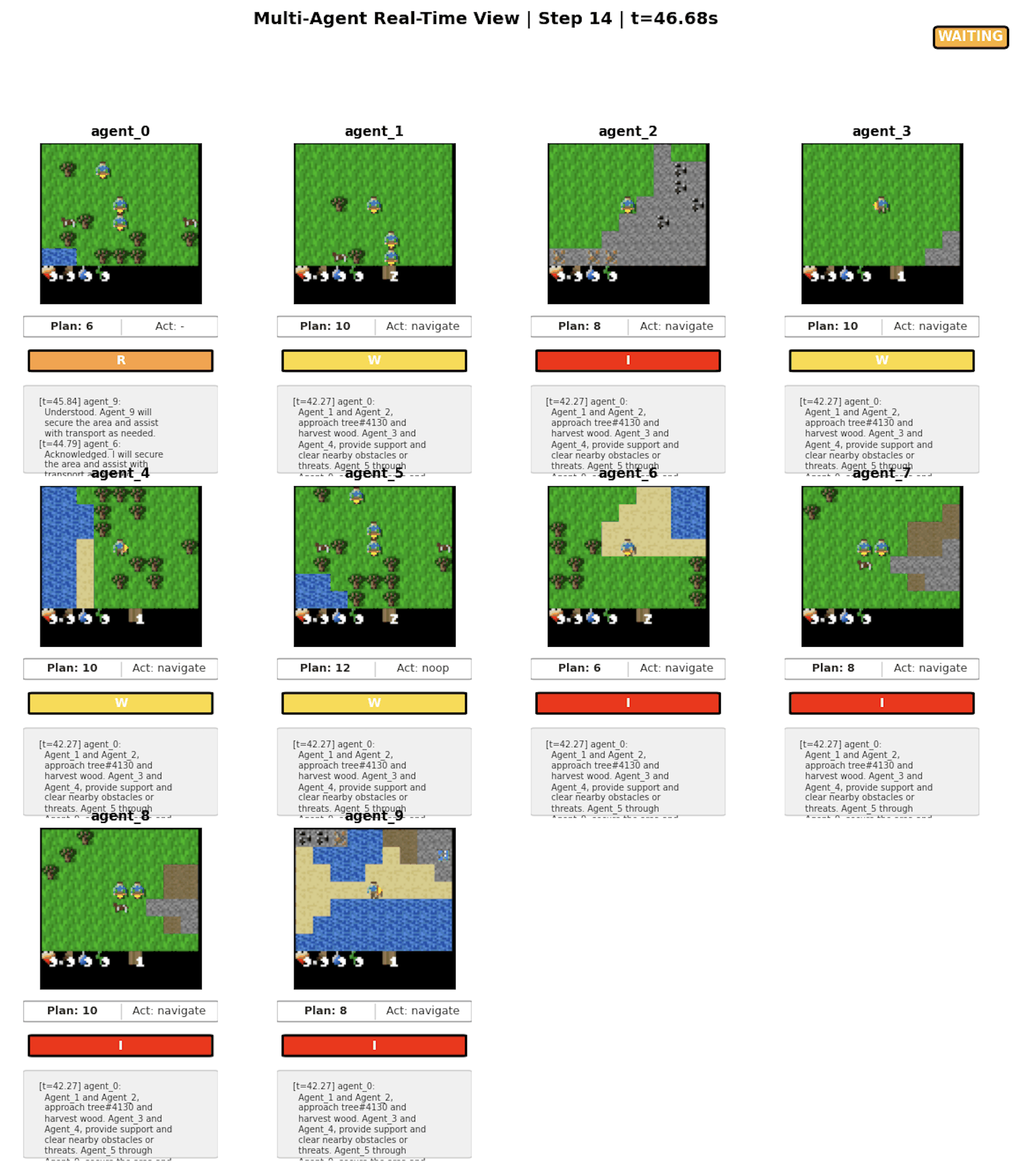}
    \caption{10-agent running example.}
    \label{fig:10-agents}
\end{figure}

COOP$^2$ is designed to scale to larger multi-agent settings. Its cooperation formalism specifies task progress through participation, capability, spatial, temporal, and dependency constraints. In MA-Crafter, the same cooperative task design can be instantiated with larger agent populations by varying participation constraints; similarly, in CUBE, task difficulty can be scaled by increasing block weights. Figure~\ref{fig:10-agents} shows a 10-agent MA-Crafter run under COOP$^2$'s MAEIL, showing that the framework can support larger teams.

\begin{figure}[H]
    \centering
    \includegraphics[width=0.8\linewidth]{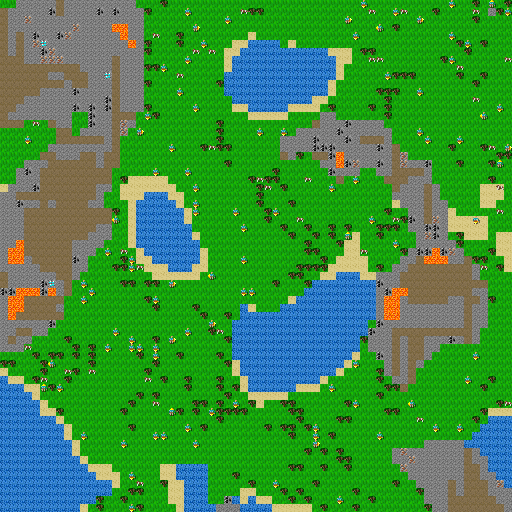}
        \caption{MA-Crafter can host 100 agents, and more!}
    \label{fig:placeholder}
\end{figure}

\begin{figure}[H]
    \centering
    \includegraphics[width=1\linewidth]{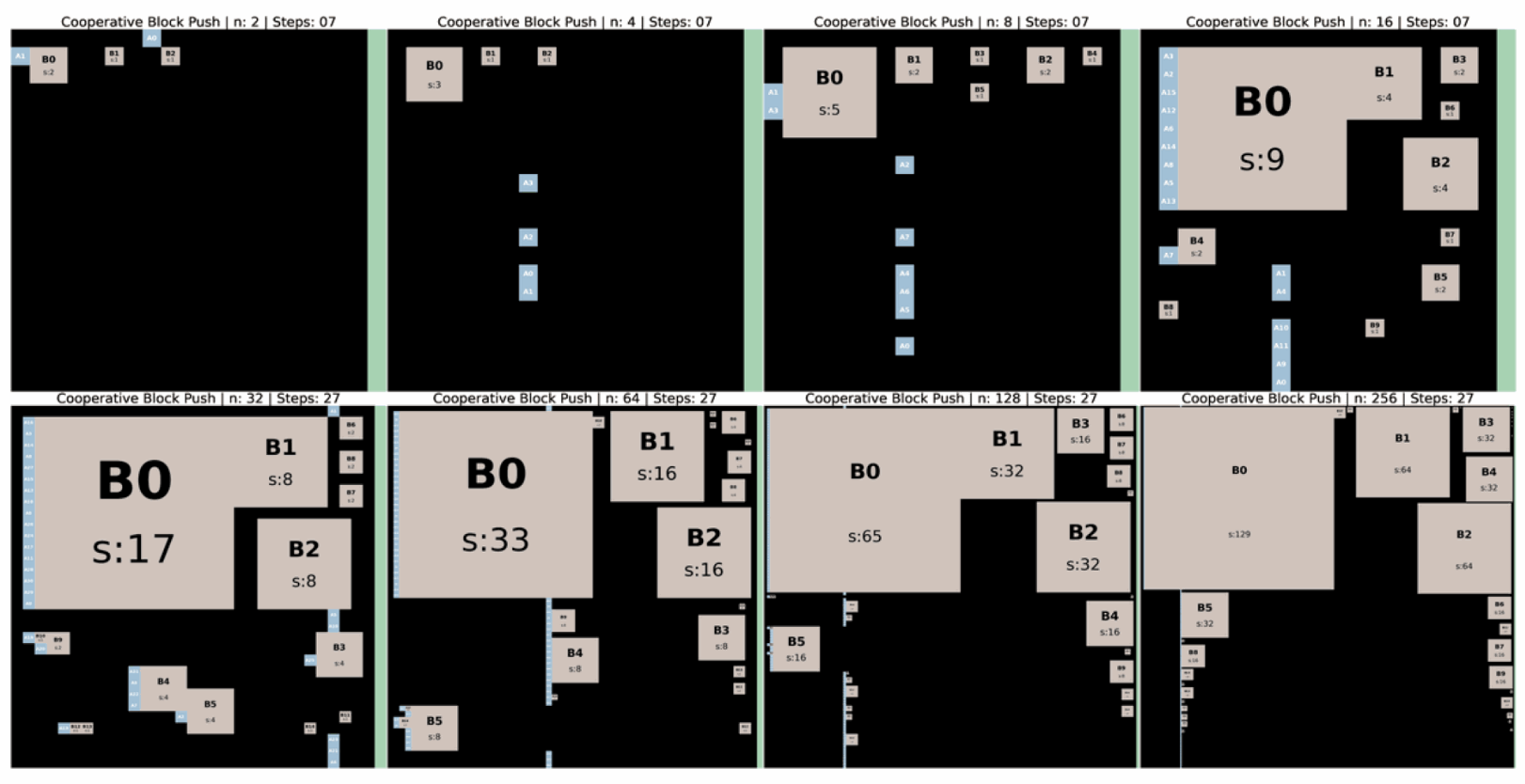}
    \caption{CUBE is a grid world where teams push weighted blocks into a goal zone while respecting embodied constraints. A single scaling parameter n jointly sets team size, block weights, and grid size, creating a transparent curriculum from small to large-scale cooperation. Each panel illustrates a snapshot of a cooperative block pushing scenario at increasing scales (n from 2 to 256).}
    \label{fig:placeholder}
\end{figure}

\subsection{Cooperation Analysis: Algorithmic Details}\label{app:env-impl}

Algorithm~\ref{alg:coop_framework} specifies the threaded execution loop used to support cooperation analysis under asynchronous communication. At each environment step, agents run in parallel within live threads so they can continuously receive and process messages while forming plans. The system waits until all agents report readiness, which serves as a synchronization barrier that ends the current reasoning interval. Once all agents are ready, their threads are joined to ensure planning has completed, and the environment is advanced via a single joint step that returns observations, rewards, and termination signals. The episode ends when any agent terminates or is truncated, or when the maximum step budget is reached. Throughout this loop, the reasoning-phase interactions (e.g., internal state updates and message traffic) can be logged during the asynchronous period, and per-step environment outputs can be recorded after each call to \texttt{env.step()} for downstream cooperation analysis.

\begin{algorithm}[H]
\caption{Threaded Multi-Agent Environment Loop}
\label{alg:coop_framework}

\KwIn{Base environment $\texttt{env\_base}$, agent name list $\texttt{agent\_names}$, maximum steps $\texttt{max\_steps}$}
\KwOut{$\texttt{last\_step}$}

\textit{Wrap base environment with \textsc{COOP2}:}\;
$\texttt{env} \leftarrow \texttt{COOP2}(\texttt{env\_base}, \texttt{agent\_names})$\;

Initialize $\texttt{done} \leftarrow \texttt{False}$\;

\While{$\texttt{env.current\_step} < \texttt{max\_steps}$ \KwSty{and} $\texttt{not done}$}{
    \textit{Run agents asynchronously to allow continuous communication}\;

    \While{not all agents are ready}{
        Ensure each agent is running in a live thread\;
        $\texttt{env.wait\_for\_state\_change(timeout=0.05)}$\;
    }

    \textit{Synchronize agents before stepping environment}\;

    Join all agent threads\;

    Step environment\;
    $(\texttt{obs}, \texttt{rewards}, \texttt{terminated}, \texttt{truncated}, \texttt{info})
    \leftarrow \texttt{env.step()}$\;

    \textit{Check termination conditions}\;
    \If{any agent terminated or truncated}{
        $\texttt{done} \leftarrow \texttt{True}$\;
    }
}

$\texttt{last\_step} \leftarrow \texttt{env.current\_step}$\;
\Return{$\texttt{last\_step}$}

\end{algorithm}

%-----

% \newpage
\subsection{Communication Structures}
\label{app:comm_struc}

% \begin{figure}[h]
%     \centering
%     \includegraphics[width=.9\linewidth]{figs/t4.png}
%     \caption{Communication topologies considered in this work}
%     \label{fig:communocation_topology}
% \end{figure}

%\graphicspath{{figures-topo/}}
To study how communication topology affects cooperative performance, we consider multiple communication topologies and analyze their impact. Nodes correspond to \textit{agents}, and a directed edge indicates that one agent can send messages to another (with reciprocity in the bidirectional case). We consider the following communication topologies:

\begin{enumerate}
    \item \textbf{Individual}: Agents do not communicate with one another.
    
    % \item \textbf{Chain}: During reasoning, each agent sends a message to the next agent in the chain. Each agent observes only the message from its immediate predecessor.
    
    \item 	\textbf{Chain}: During reasoning, each agent broadcasts messages to all other agents. Each agent observes all previously generated messages.
    
    \item \textbf{Centralized}: For n agents, one agent acts as a leader and the remaining agents are followers. During reasoning, the leader broadcasts a message to all followers and waits for their responses. Each follower waits for the leader’s message, generates a response, and forms a plan. If a follower is interrupted by a new leader message during execution, it decides whether to resume the current plan or replan before responding to the leader.
    
    % \item \textbf{Hierarchical}: Each agent has at most one parent and an arbitrary number of children. During reasoning, an agent sends messages to its immediate children and waits for their responses.
    % \item \textbf{Decentralized}: At each decision-making step, each agent has a communication budget of n. An agent may send up to n messages and receive up to n messages per step.
\end{enumerate}

% One knob to tune figure height
\newcommand{\Topotimeheight}{3.5cm}

\begin{figure}[H]
  \centering

  \begin{subfigure}[c]{0.48\textwidth}
    \centering
    \includegraphics[height=\Topotimeheight]{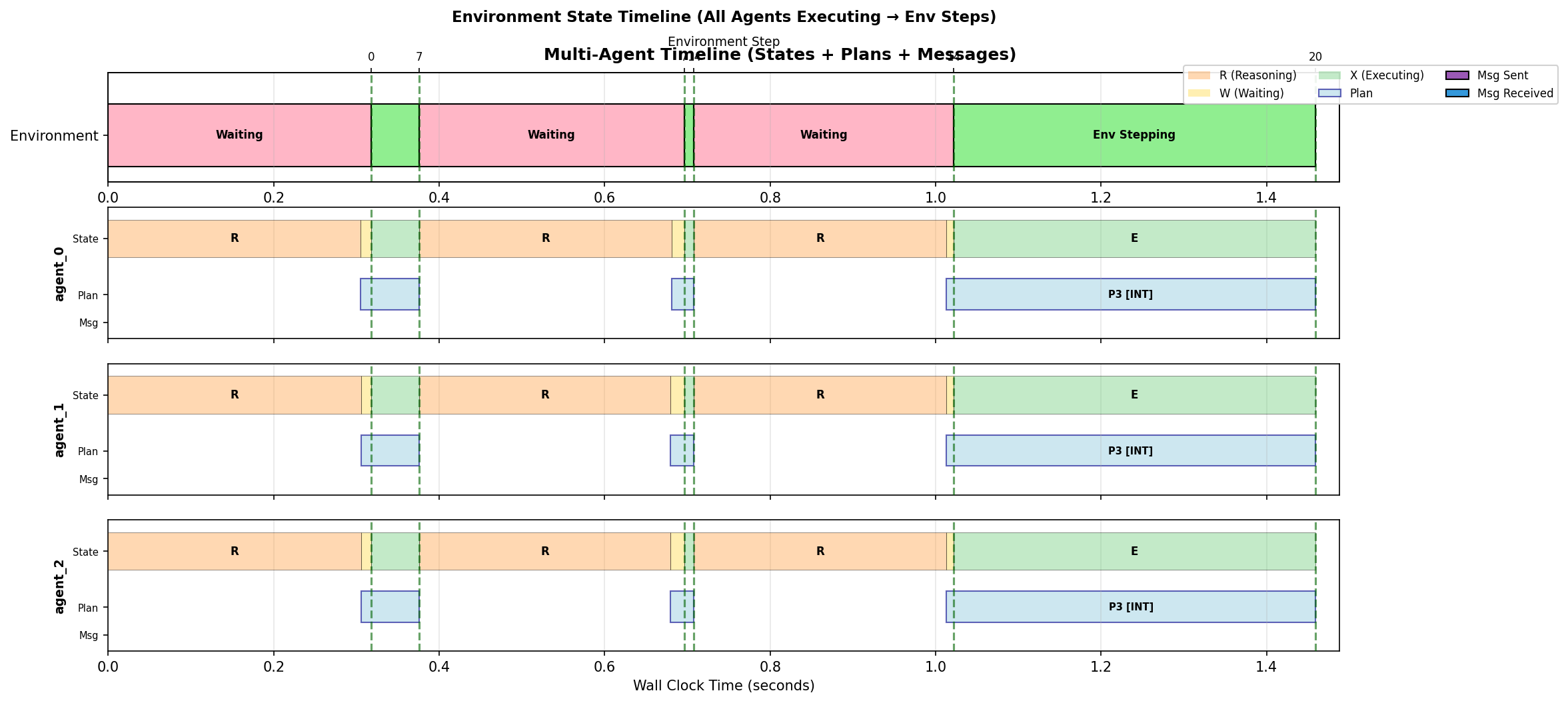}
    \caption{\textbf{Individual baseline.} No communication: each agent plans independently without waiting or sending messages.}
    \label{fig:topo_individual}
  \end{subfigure}
  \hfill
  \begin{subfigure}[c]{0.48\textwidth}
    \centering
    \includegraphics[height=\Topotimeheight]{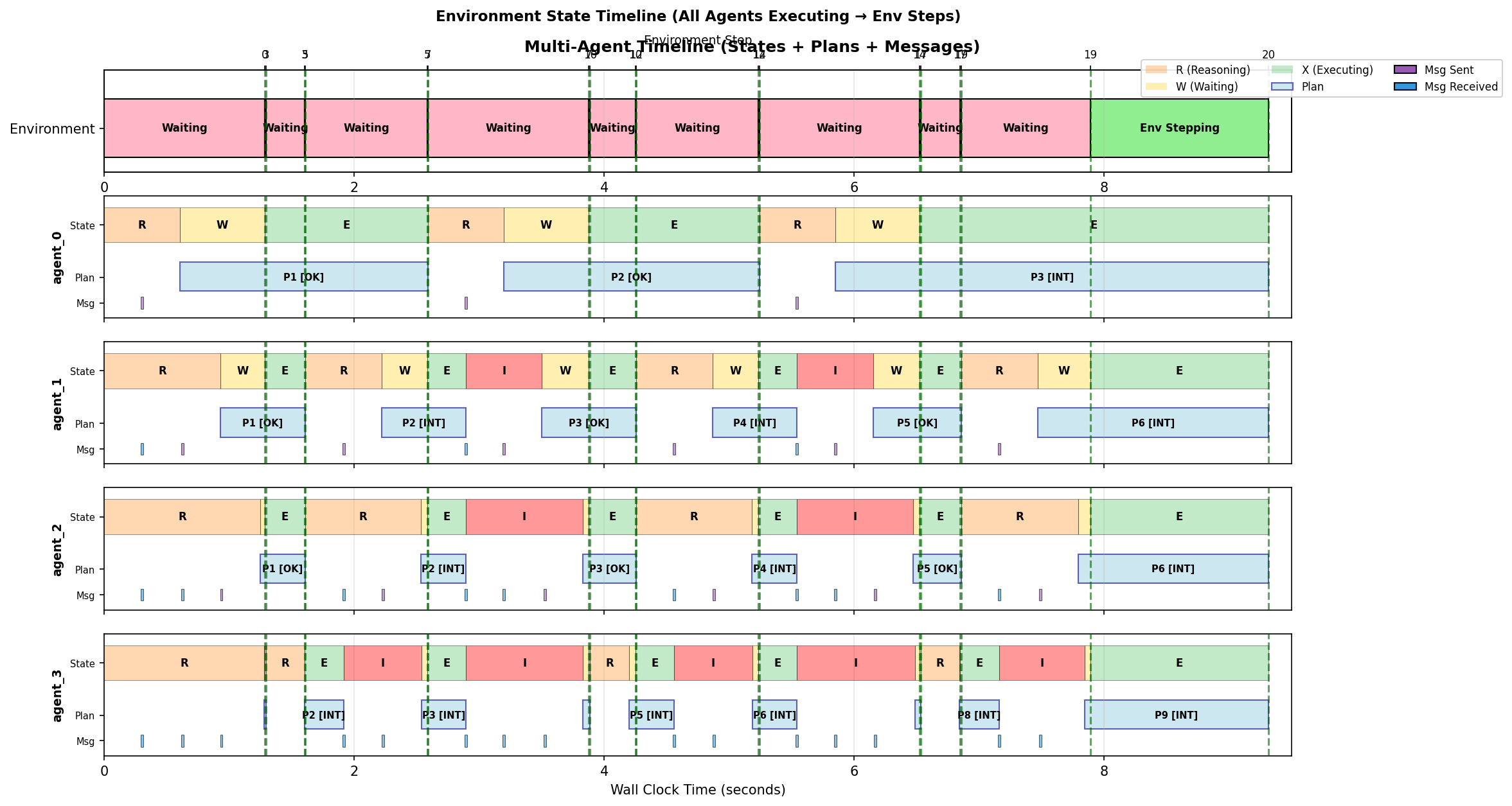}
    \caption{\textbf{Chain topology.} Agents speak in order: each waits for the previous speaker, then broadcasts to all following agents before planning.}
    \label{fig:topo_debate}
  \end{subfigure}

  \begin{subfigure}[c]{0.48\textwidth}
    \centering
    \includegraphics[height=\Topotimeheight]{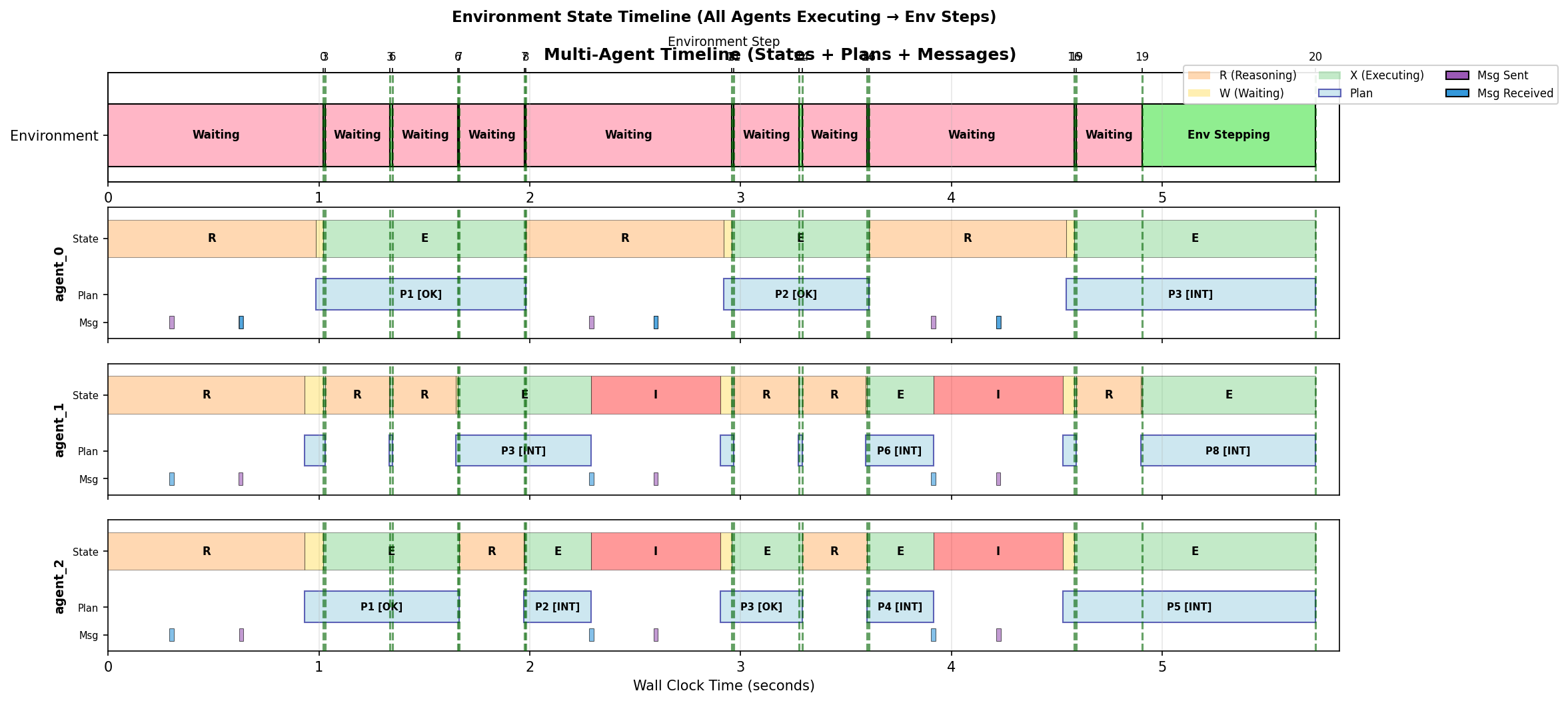}
    \caption{\textbf{Centralized topology.} Leader broadcasts to all followers, waits for their responses, then plans; followers acknowledge the leader and plan.}
    \label{fig:topo_centralized}
  \end{subfigure}
  \hfill
  % \begin{subfigure}[c]{0.48\textwidth}
  %   \centering
  %   \includegraphics[height=\Topotimeheight]{decentralized/comprehensive_timeline.png}
  %   \caption{\textbf{Decentralized topology.} Free communication with bounded per-step send and receive budgets; agents process a limited number of messages and optionally message peers before planning.}
  %   \label{fig:topo_decentralized}
  % \end{subfigure}

  \caption{Timeline visualizations for the three communication structures considered in this work.}
  \label{fig:topo_timelines}
\end{figure}

\subsection{Communication Structure Prompt}
Each communication structure includes role-specific instructions that are appended to the agent's system prompt. The role descriptions below define how each agent interprets its position in the team under Individual, Centralized, and Chain settings.
\subsubsection{Centralized} \label{prompt:centralized}
\begin{tcolorbox}[breakable, title=Role Descriptions in Centralized Topology, colback=gray!2, colframe=black!60, fonttitle=\bfseries]
\begin{lstlisting}[style=jsonstyle]
{
    LEADER_ROLE = """
        ## Your Role: LEADER
        You are the leader of a team. Your responsibilities:
        - Coordinate team activities by broadcasting directives to followers
        - Wait for acknowledgments from all followers before finalizing your plan
        - Make strategic decisions that benefit the whole team
    """
    
    FOLLOWER_ROLE = """
        ## Your Role: FOLLOWER
        You are a follower in a team. Your responsibilities:
        - Wait for and follow the leader's directives
        - Acknowledge the leader's messages
        - Execute tasks that support the team's goals
    """
}
\end{lstlisting}
\end{tcolorbox}

\subsubsection{Chain} \label{prompt:debate}
\begin{tcolorbox}[breakable, title=Role Descriptions in Chain Topology, colback=gray!2, colframe=black!60, fonttitle=\bfseries]
\begin{lstlisting}[style=jsonstyle]
{
    if speaker_order == 0:
        return """
        ## Your Role: FIRST SPEAKER
            You speak first in the chain. Your responsibilities:
            - Set the initial direction and strategy for the team
            - Broadcast your analysis to all other agents
            - Your message will be heard by everyone
        """
    elif speaker_order == n_agents - 1:
        return """
        ## Your Role: FINAL SPEAKER
            You speak last in the chain. Your responsibilities:
            - Consider all previous speakers' messages
            - Make a final decision incorporating all viewpoints
            - You have heard from everyone else
        """
    else:
        return f"""
        ## Your Role: SPEAKER {speaker_order + 1} of {n_agents}
            You speak in the middle of the chain. Your responsibilities:
            - Listen to the previous speaker
            - Add your perspective and broadcast to remaining agents
            - Build on the discussion so far
        """
}
\end{lstlisting}
\end{tcolorbox}

% \subsubsection{Decentralized} \label{prompt:decentralized}
% \begin{tcolorbox}[breakable, title=Role Descriptions in Decentralized Topology, colback=gray!2, colframe=black!60, fonttitle=\bfseries]
% \begin{lstlisting}[style=jsonstyle]
% {
%     DECENTRALIZED_ROLE = """
%     ## Your Role: DECENTRALIZED AGENT
%         You operate in a decentralized network with free communication. Your responsibilities:
%         - Communicate with any other agents as needed
%         - Process incoming messages from peers
%         - Make independent decisions while coordinating with others
%         - Balance between communication and task execution
%     """
% }
% \end{lstlisting}
% \end{tcolorbox}

\subsubsection{Individual} \label{prompt:individual}
\begin{tcolorbox}[breakable, title=Role Descriptions in Individual Topology, colback=gray!2, colframe=black!60, fonttitle=\bfseries]
\begin{lstlisting}[style=jsonstyle]
{
    INDIVIDUAL_ROLE = """
        ## Your Role: INDEPENDENT AGENT
        You operate independently without communication with other agents.
        Make decisions based solely on your own observations and goals.
    """
}
\end{lstlisting}
\end{tcolorbox}

%---

 %cluster 2: imp detail, algo, prompt
\newpage
\newcommand{\ind}{\mathbf{I}}

\section{Environments}
This cluster details the two cooperative environments used in our experiments, MA-Crafter and CUBE, including their task structures, action spaces, observation modalities, and the prompts presented to LLM agents.

\subsection{MA-Crafter}
\label{app:ma-crafter}

MA-Crafter is an embodied multi-agent environment that extends the Crafter world to support cooperative decision making among many agents in a shared, partially observable 2D grid world. Agents explore, gather resources, and craft tools through a technology tree, while maintaining individual inventories, health states, and local views. We apply Coop$^2$’s task design to instantiate scalable cooperative tasks by enforcing embodied participation constraints on execution, and provide a dual interface: a symbolic planning layer for high-level, interpretable multi-agent plans, and a low-level execution layer that grounds plan executions under these constraints.

\begin{figure}[H]
    \centering
    \includegraphics[width=1\linewidth]{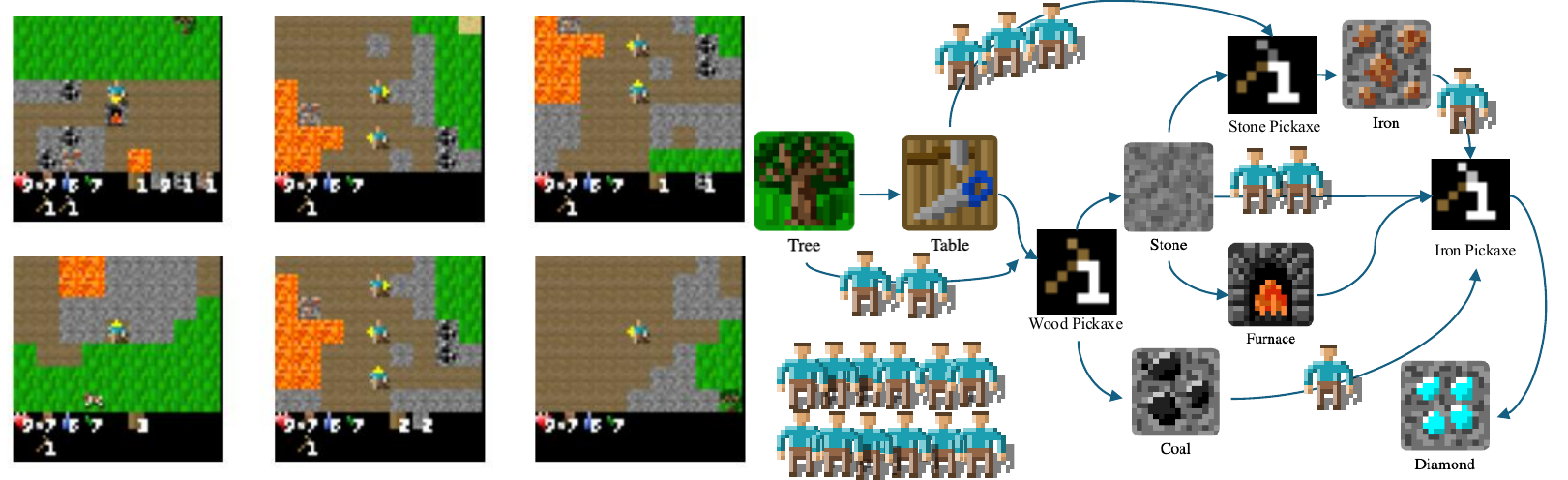}
    \caption{MA-Crafter}

        \label{fig:placeholder}
\end{figure}

\begin{figure}[H]
    \centering
    \includegraphics[width=1\linewidth]{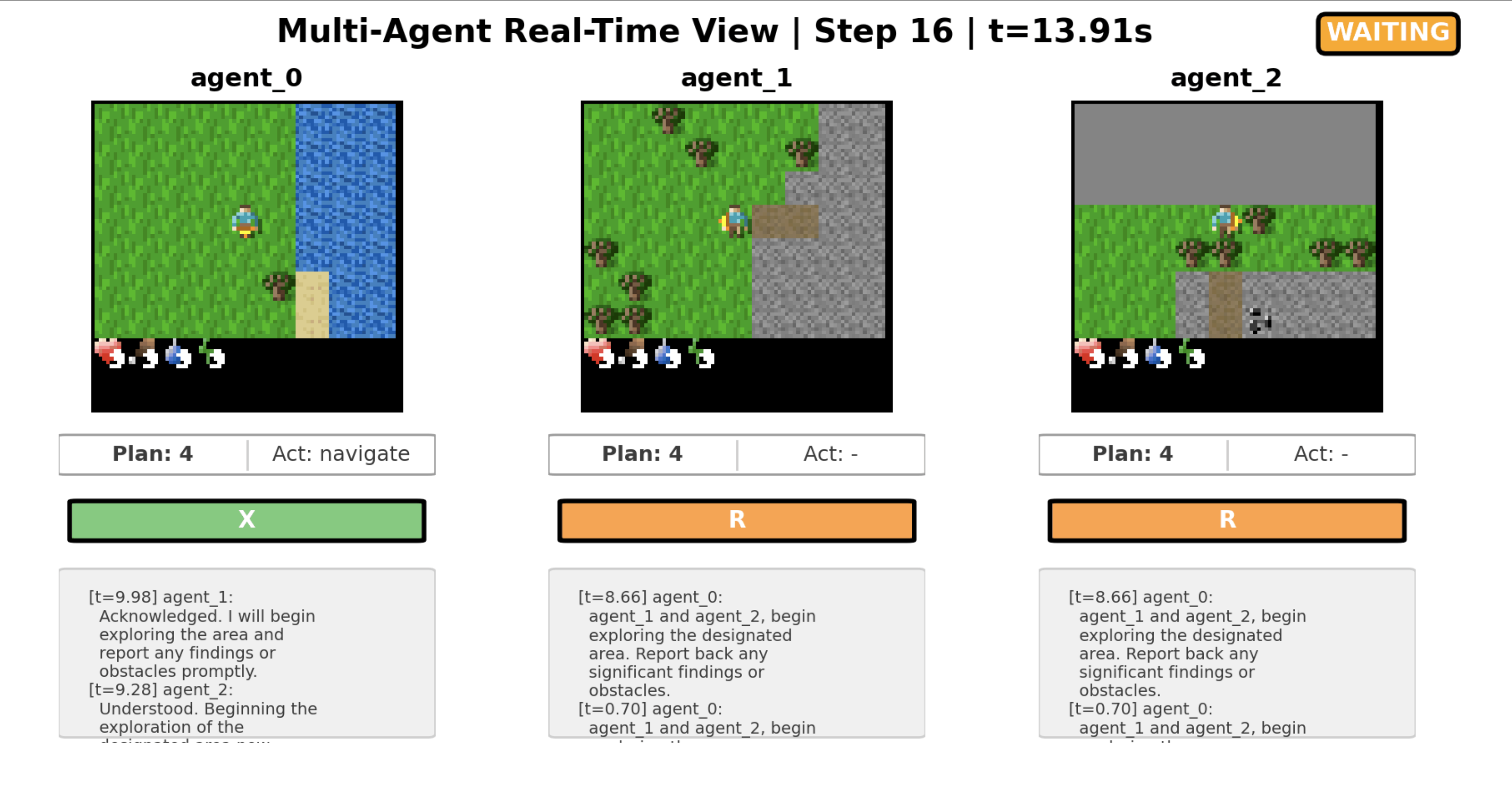}
    \caption{MA-Crafter Experiment}
    \label{fig:placeholder}
\end{figure}

\subsubsection{Actions}
\begin{table}[h]
\scriptsize
\renewcommand{\arraystretch}{1.2}
\setlength{\tabcolsep}{6pt}
\centering
\begin{tabularx}{\linewidth}{@{}l m{0.22\linewidth} m{0.60\linewidth}@{}}
\toprule
\textbf{Action} & \textbf{Arguments} & \textbf{Effect} \\
\midrule
\texttt{noop}     & (\textit{num\_steps})             & Do nothing for a specified number of steps (default: 1). \\
\texttt{move}     & (\textit{direction}, \textit{num\_steps}) & Move in a direction (left, right, up, down) for a specified number of steps (default: 1). \\
\texttt{collect}  & (\textit{Resource})                             & Collect a resource in front of the agent. All participating agents receive the resource if participation requirements are met. \\
\texttt{craft}    & (\textit{object\_type})          & Craft an item (e.g., wood\_pickaxe, stone\_sword). All participating agents receive the crafted item if participation requirements are met. \\
\texttt{sleep}    & ( )                             & Sleep to recover energy. \\
\texttt{place}    & (\textit{object\_type})          & Place an object in the environment (e.g., stone, table, furnace, plant). \\
\texttt{share}    & (\textit{recipient\_agent\_id}, \textit{resource\_type}, \textit{quantity}) & Share resources/tools with another agent. Quantity defaults to 1. \\
\texttt{navigate} & (\textit{object\_type}, \textit{item\_id}, \textit{timeout}) & Navigate to a specific object using pathfinding. Multi-step action that completes when agent reaches target or times out (default timeout: 64 steps). \\
\bottomrule
\end{tabularx}
\caption{Symbolic actions in MCrafter used in our experiments. Action space covers all high-level behaviors available to agents.}
\label{tab:symb_actions}
\end{table}

Although the symbolic action set is compact, these actions remain expressive because coordination emerges through agents’ choices of where to move, when to wait, and which block to push, all under embodied constraints from collisions, alignment, and quorum requirements.

\begin{lstlisting}[style=pythonstyle, caption={Pydantic schema for LLM response}, label={lst:primitive_action}]

class Direction(str, Enum):
    """Movement directions."""
    UP = "up"
    DOWN = "down"
    LEFT = "left"
    RIGHT = "right"


class Resource(str, Enum):
    """Collectible resources."""
    WOOD = "wood"
    STONE = "stone"
    COAL = "coal"
    IRON = "iron"
    DIAMOND = "diamond"


class PlaceableItem(str, Enum):
    """Items that can be placed."""
    TABLE = "table"
    FURNACE = "furnace"
    PLANT = "plant"


class CraftableItem(str, Enum):
    """Items that can be crafted."""
    WOOD_PICKAXE = "wood_pickaxe"
    STONE_PICKAXE = "stone_pickaxe"
    IRON_PICKAXE = "iron_pickaxe"
    WOOD_SWORD = "wood_sword"
    STONE_SWORD = "stone_sword"
    IRON_SWORD = "iron_sword"


class Task(str, Enum):
    """Available tasks/achievements in the environment."""
    # Collection tasks
    COLLECT_WOOD = "collect_wood"
    COLLECT_STONE = "collect_stone"
    COLLECT_COAL = "collect_coal"
    COLLECT_IRON = "collect_iron"
    COLLECT_DIAMOND = "collect_diamond"
    COLLECT_DRINK = "collect_drink"
    COLLECT_SAPLING = "collect_sapling"
    
    # Crafting tasks
    MAKE_WOOD_PICKAXE = "make_wood_pickaxe"
    MAKE_STONE_PICKAXE = "make_stone_pickaxe"
    MAKE_IRON_PICKAXE = "make_iron_pickaxe"
    MAKE_WOOD_SWORD = "make_wood_sword"
    MAKE_STONE_SWORD = "make_stone_sword"
    MAKE_IRON_SWORD = "make_iron_sword"
    
    # Placement tasks
    PLACE_TABLE = "place_table"
    PLACE_FURNACE = "place_furnace"
    PLACE_PLANT = "place_plant"
    PLACE_STONE = "place_stone"


class TaskSpecification(BaseModel):
    """Task specification with object type and target ID."""
    task: Task = Field(description="The task type to accomplish")
    object_type: str = Field(description="Type of target object (e.g., 'tree', 'stone', 'cow')")
    object_id: int = Field(description="ID of the specific object instance to target")
    
    def __str__(self) -> str:
        """String representation of the task specification."""
        return f"{self.task.value}({self.object_type}#{self.object_id})"


class InterruptDecision(str, Enum):
    """Decision options when agent is interrupted."""
    RESUME = "resume"  # Continue with the current plan
    REPLAN = "replan"  # Generate a completely new plan


# ============================================================================
# Pydantic models for structured LLM output - Action Classes
# ============================================================================

class MoveAction(BaseModel):
    """Move in a direction."""
    action_type: Literal["move"] = "move"
    direction: Direction = Field(description="Direction to move")
    num_steps: int = Field(ge=1, le=5, description="Number of steps (1-5)")


class CollectAction(BaseModel):
    """Collect a resource at current location."""
    action_type: Literal["collect"] = "collect"
    target: Resource = Field(description="Resource to collect")


class PlaceAction(BaseModel):
    """Place an item in the world."""
    action_type: Literal["place"] = "place"
    item: PlaceableItem = Field(description="Item to place")


class CraftAction(BaseModel):
    """Craft an item."""
    action_type: Literal["craft"] = "craft"
    item: CraftableItem = Field(description="Item to craft")


class SleepAction(BaseModel):
    """Sleep to restore energy."""
    action_type: Literal["sleep"] = "sleep"


class NoopAction(BaseModel):
    """Do nothing this step."""
    action_type: Literal["noop"] = "noop"


class NavigateAction(BaseModel):
    """Navigate to a specific object in the world using pathfinding."""
    action_type: Literal["navigate"] = "navigate"
    object_type: str = Field(description="Type of object to navigate to (e.g., 'tree', 'stone', 'coal', 'iron', 'diamond', 'water', 'table', 'furnace')")
    item_id: int = Field(description="ID of the specific object instance to navigate to")
    timeout: int = Field(default=30, ge=1, le=100, description="Maximum steps to attempt navigation (default 30)")


class ShareAction(BaseModel):
    """Share resources with another agent."""
    action_type: Literal["share"] = "share"
    recipient_agent_id: str = Field(description="Agent ID to share with (e.g., 'agent_0', 'agent_1')")
    resource_type: str = Field(description="Type of resource to share (e.g., 'wood', 'stone', 'coal', 'iron')")
    quantity: int = Field(default=1, ge=1, description="Amount to share (default 1)")


# Union type for all actions
LLMAction = Union[MoveAction, CollectAction, PlaceAction, CraftAction, SleepAction, NoopAction, NavigateAction, ShareAction]


# ============================================================================
# Response Models
# ============================================================================

class LLMPlanResponse(BaseModel):
    """Structured response from LLM for plan generation."""
    task: TaskSpecification = Field(description="The task specification with type and optional target object")
    actions: List[LLMAction] = Field(description="List of actions to execute")
    reasoning: str = Field(description="Brief explanation of why this plan was chosen")


class LLMMessageResponse(BaseModel):
    """Structured response from LLM for message generation."""
    recipients: List[str] = Field(description="List of agent IDs to send message to")
    content: str = Field(description="Message content to send")
    reasoning: str = Field(description="Brief explanation of why sending this message")


class LLMInterruptResponse(BaseModel):
    """Structured response from LLM for interrupt handling."""
    decision: InterruptDecision = Field(
        description="Whether to resume the current plan or generate a new plan"
    )
    reasoning: str = Field(
        description="Brief explanation of why this decision was made based on the messages received"
    )
    # Optional new plan - only required if decision is REPLAN
    new_plan: Optional[LLMPlanResponse] = Field(
        default=None,
        description="The new plan to execute (required if decision is 'replan')"
    )

\end{lstlisting}

\subsubsection{Task Design}

We apply cooperative task design to resource collection in MA-Crafter by defining each collectible resource as a task $g\in\mathcal{G}$, where completion requires $p(g)$ agents to \texttt{collect} simultaneously within radius $d$ of $g$ and with the required tools.

\begin{gather}
C_{\mathrm{n}}^{i}(g,t) = \mathrm{cap}_i(t)\succeq \mathrm{cap}(g,t), \\
C_{\mathrm{\Delta n}}^{i}(g,t;d) = \lVert x_{i,t}-y_g\rVert \le d, \\
C_{\mathrm{\Delta t}}^{i}(g,t) = a_{i,t}=\mathrm{act}(g), \\
N(C)(g,t) = \sum_{i\in I}\ind\!\left[C^{i}(g,t)\right] \\
C_{\mathrm{n}}(g,t;d,p(g)) = \ind\!\left[
\bigwedge_{x\in\{\Delta \ell, \Delta t, n\}}
N(C_x)(g,t)\ge p(g)
\right]
\end{gather}

\subsection{COOP$^2$ Task Design in \texttt{MA-Crafter}}
\label{sec:mac-difficulty_design}

\begin{table}[H]
\centering
\small
\begin{tabular}{lcc}
\toprule
Resource & Required agents & Required tool \\
\midrule
tree / wood & 1 & -- \\
stone       & 1 & \texttt{wood\_pickaxe} \\
coal        & 2 & \texttt{wood\_pickaxe} \\
iron        & 3 & \texttt{stone\_pickaxe} \\
diamond     & 3 & \texttt{iron\_pickaxe} \\
cow         & 1 & -- \\
plant       & 1 & -- \\
\bottomrule
\end{tabular}
\caption{MA-Crafter cooperative collection configuration. The spatial
distance threshold is \(1\) Manhattan step for all resources.}
\label{tab:macrafter-coop-config}
\end{table}

\begin{table}[H]
\centering
\small
\begin{tabular}{lc}
\toprule
Scored resource & Base value \\
\midrule
wood    & 1 \\
stone   & 10 \\
coal    & 25 \\
iron    & 25 \\
diamond & 80 \\
\bottomrule
\end{tabular}
\caption{MA-Crafter team-score values.}
\label{tab:macrafter-score-config}
\end{table}

\subsubsection{Environment Prompt}

The environment prompt is a fixed textual description that outlines the basic rules, actions, and constraints of the \texttt{MA-Crafter} world. It is provided to all agents at the beginning of each planning step and serves as a reference for understanding the world. While it describes prerequisites, the actual environment configuration is dynamically defined in \texttt{env\_rule\_txt} (see Section~\ref{sec:mac-difficulty_design}) and passed to agents at each step. These parameters-including object weights, required collaborators, and crafting constraints-may be redefined over time, enabling emergent adaptability challenges in multi-agent systems.\\

\begin{tcolorbox}[breakable, colback=gray!2, colframe=black!60, title=\texttt{MA-Crafter} Environment Specification, fonttitle=\bfseries]
    \label{prompt:env}
    \textbf{Goal} \\
    The team's objective is to \textbf{collect a diamond}. Any agent on the team successfully collecting a diamond achieves the goal. \\
    
    \textbf{Cooperative Configuration} \\
    The environment uses a configuration file that defines cooperative requirements for each resource, provided in observations as ``COOPERATIVE CONFIGURATION''. Key parameters: \\
    \hspace*{1em}-- Distance threshold: Maximum distance (in tiles) for an agent to be considered ``near'' a resource. \\
    \hspace*{1em}-- Resource requirements: For each resource type, \texttt{required\_agents} (minimum agents needed, all must be within distance and issue collect action) and \texttt{required\_tool} (tool that ALL participating agents must have). \\
    \hspace*{1em}-- Different levels have different requirements: Level 1 (most resources require 1 agent), Level 2 (most require 2 agents), Level 3 (varying numbers up to 4 agents). \\
    
    \textbf{Actions} \\
    \texttt{noop}: Does nothing for the current step. Can be used with \texttt{num\_steps} parameter to wait/synchronize with other agents. \\
    \texttt{move}: Moves the agent one step in the specified direction. Use to explore when object IDs are unknown. \\
    \texttt{navigate}: Multi-step action that moves the agent to reach a specified target by ID. The agent ends up facing the target. If the target is blocked, moves to the nearest location. Fails if object type and ID mismatch or object ID does not exist. \\
    \texttt{sleep}: Multi-step action that puts the agent to sleep until energy is fully recovered. \\
    \texttt{place}: Places an object of the specified type into the environment where the agent is facing. \\
    \texttt{craft}: Creates a new item using resources from the leader's bag. \\
    \hspace*{1em}-- The leader must be facing the required tool station. \\
    \hspace*{1em}-- At least \texttt{x} agents (including the leader) must be near the tool station. \\
    \hspace*{1em}-- All agents must execute the \texttt{craft} action simultaneously. \\
    \hspace*{1em}-- Only the leader's bag is used for resource consumption. \\
    \texttt{collect}: Collects a resource from the environment. \\
    \hspace*{1em}-- Agent MUST be nearby the resource (use \texttt{navigate} to get close first). \\
    \hspace*{1em}-- At least \texttt{x} agents must be within distance \texttt{d} of the object (check COOPERATIVE CONFIGURATION). \\
    \hspace*{1em}-- All participating agents must perform the \texttt{collect} action simultaneously. \\
    \hspace*{1em}-- The resource is added to participating agents' bags (if space allows). \\
    \hspace*{1em}-- Usage: (1) If object ID known: use \texttt{navigate(object\_type, object\_id)}; (2) If unknown: use \texttt{move} to explore; (3) Once nearby, use \texttt{collect}. \\
    \texttt{share}: Transfers resources from your inventory to another agent. \\
    \hspace*{1em}-- Parameters: \texttt{recipient\_agent\_id} (e.g., ``agent\_0''), \texttt{resource\_type} (e.g., ``wood'', ``stone''), \texttt{quantity} (amount to transfer). \\
    \hspace*{1em}-- Conditions: You must have the resource; cannot share health; cannot share with yourself. \\
    
    \textbf{Bag Capacity} \\
    Each resource has a weight; the total carried weight must not exceed the bag capacity when collecting or sharing. \\
    
    \textbf{Prerequisites} \\
    \textit{Collecting Resources}: \\
    - \texttt{cow}: navigate to cow, then collect (restores food) \\
    - \texttt{drink (water)}: navigate to water, then collect (restores drink) \\
    - \texttt{wood}: navigate to tree, then collect (no tool required) \\
    - \texttt{stone}: navigate to stone, then collect (requires \texttt{wood\_pickaxe}) \\
    - \texttt{coal}: navigate to coal, then collect (requires \texttt{wood\_pickaxe}) \\
    - \texttt{iron}: navigate to iron, then collect (requires \texttt{stone\_pickaxe}) \\
    - \texttt{diamond}: navigate to diamond, then collect (requires \texttt{iron\_pickaxe}) \\
    \textit{Placing Structures}: \\
    - \texttt{table}: consumes 2 wood; must be placed on grass \\
    - \texttt{furnace}: consumes 4 stone \\
    \textit{Crafting Tools} (must be near required station): \\
    - \texttt{wood\_pickaxe}: crafted at a \texttt{table}; uses 1 wood \\
    - \texttt{stone\_pickaxe}: crafted at a \texttt{table}; uses 1 stone and 1 wood \\
    - \texttt{iron\_pickaxe}: crafted at a \texttt{furnace}; uses 1 iron, 1 coal, and 1 wood \\
    
    \textbf{Agent Health} \\
    High numbers mean better condition. To replenish: \\
    \hspace*{1em}-- Low hunger: collect from cow \\
    \hspace*{1em}-- Low thirst: collect water \\
    \hspace*{1em}-- Low energy: sleep \\
    
    \textbf{World Rules} \\
    1. Satisfy prerequisites (tools and materials). \\
    2. Navigate to resources using \texttt{navigate}. \\
    3. Use the \texttt{collect} action once nearby. \\
    4. Be cooperative: help others when needed and ask for help when necessary. \\
    5. Your collaborators should not include yourself. \\
    6. Work with no more agents than needed. \\
    \end{tcolorbox}

\subsection{CUBE}\label{app:cube}
\begin{wrapfigure}{r}{0.25\textwidth}
    \vspace{-9pt}
    \centering
    \includegraphics[width=0.9\linewidth]{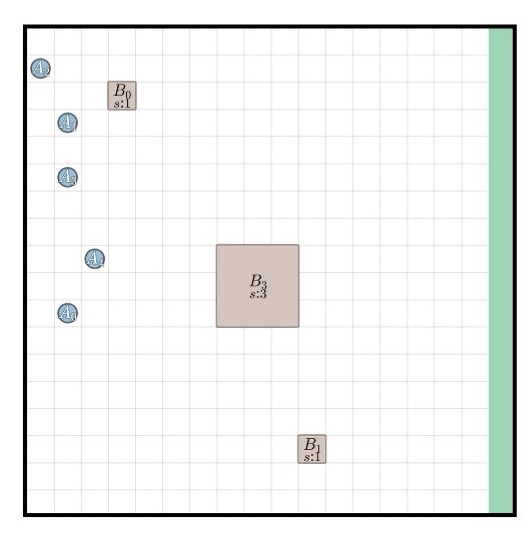}
    \caption{CUBE}
    \label{fig:placeholder}
  \end{wrapfigure}
At its base level, CUBE is a grid-world environment built on PettingZoo’s parallel API~\citep{terry2021pettingzoo} and modified from the multi-agent block-pushing environment introduced in~\citep{yangcube}. 

Compared to the original formulation, we simplify the environment design while retaining its cooperative block-pushing objective. Agents must cooperate to push square blocks into a designated goal region. The resulting dynamics create coordination challenges through collisions, congestion, and enforced cooperation. The same underlying environment has also been used in~\citep{nourzad2025well}.

The environment consists of agents and movable blocks placed on a grid. Their dynamics create coordination challenges through collisions, congestion, and enforced collaboration.

% \begin{figure}[H]
%     \centering
%     \includegraphics[width=0.3\linewidth]{figs/n5.jpg}
%     \caption{CUBE}
%     \label{fig:placeholder}
% \end{figure}

\begin{figure}
    \centering
    \includegraphics[width=0.9\linewidth]{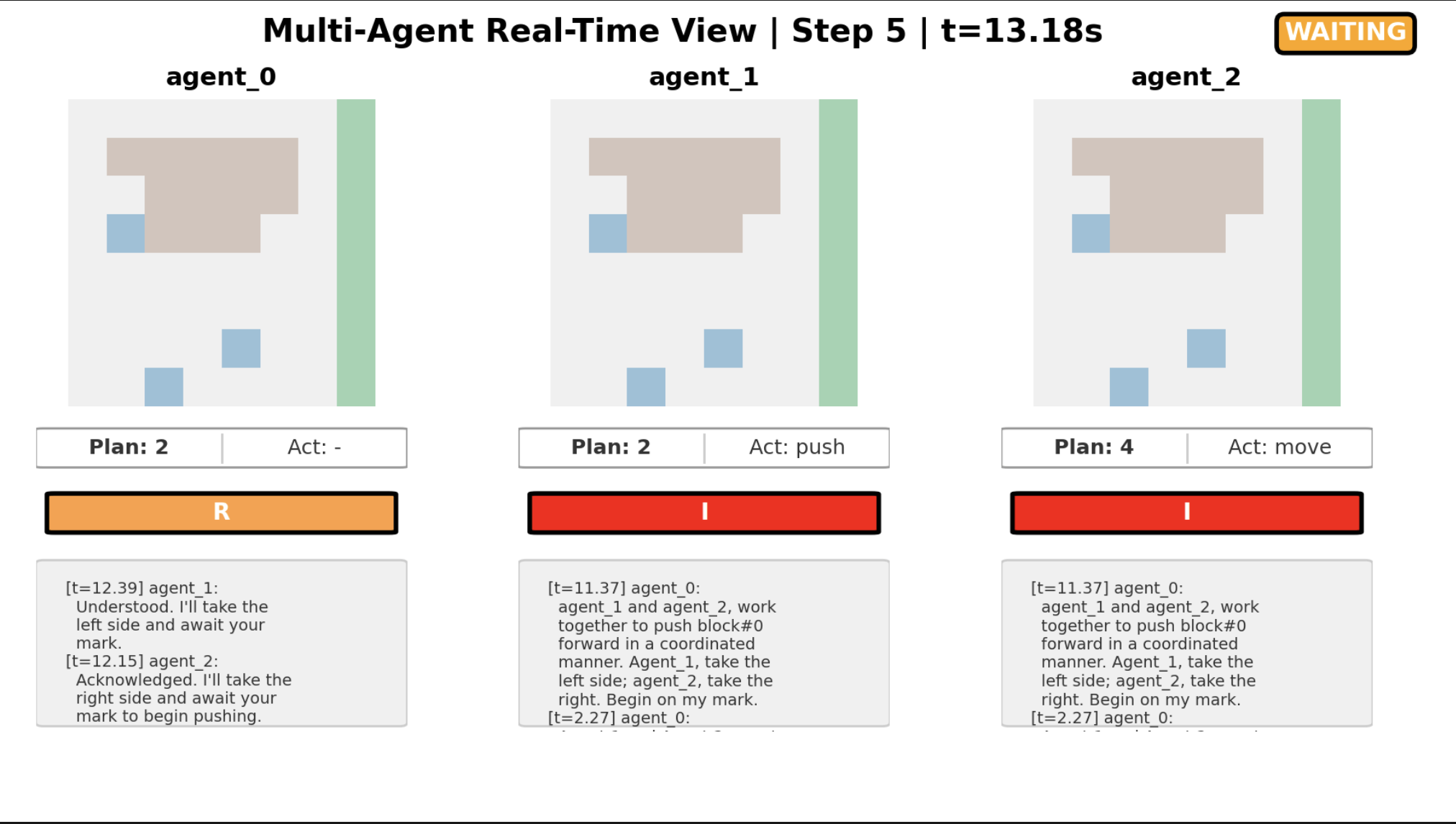}
    \caption{CUBE Experiment}
    \label{fig:placeholder}
\end{figure}

\subsubsection{Task Design}

We denote the set of agents by I and the set of blocks available at the start of an episode by $\mathcal{B}$.
We treat each block's four sides as distinct tasks, denoted by $g\in \mathcal G$, where each task corresponds to pushing block $b(g)$ from side $s(g)\in\mathcal{S}(b(g))$. Each block has an integer weight $w_{b(g)} \geq 1$, indicating the number of agents needed to push it. Meaning, each task requires participation threshold $p(g)=w_{b(g)}$.
The block occupies a contiguous square of side length $w_{b(g)}$, so its physical size grows in direct proportion to its weight.
This proportionality ensures consistency, as larger blocks both span more grid cells and require a greater quorum of agents to move. 
Each agent $i \in I$ occupies a single grid cell, with position denoted $\mathcal x_i^t$ at step $t$. Episodes \textit{terminate} successfully when all blocks have been delivered to the goal region.  
Episodes \textit{truncate} if the \texttt{max\_steps} is reached without delivering all blocks. 

\begin{gather}
C_{\mathrm{n}}^{i}(g,t) = \mathrm{cap}_i(t), \\
C_{\mathrm{\Delta \ell}}^{i}(g,t) = \min_{g\in \mathcal{G}_{g}}\lVert \mathcal x_{i,t}-g\rVert \le 1, \\
C_{\mathrm{\Delta t}}^{i}(g,t) = a_{i,t}=\texttt{push}(b(g)), \\
N(C)(g,t) = \sum_{i\in I}\ind\!\left[C^{i}(g,t)\right], \\
C_{\mathrm{n}}(g,t) = \ind\!\left[
\bigwedge_{x\in\{\Delta \ell, \Delta t, n\}}
N(C_x)(g,t)\ge w_{b(g)}
\right]
\end{gather}

\begin{table}[H]
\centering
\small
\begin{tabular}{lcc}
\toprule
Preset & Block specification\\
\midrule
Setting & 9 blocks in total: 3 blocks each of weights 1, 2, and 3 \\
\bottomrule
\end{tabular}
\caption{CUBE cooperative block-pushing configuration. A block's weight is
both its side length and the number of agents required to push it.}
\label{tab:cube-block-config}
\end{table}

\begin{wrapfigure}{r}{0.34\textwidth}
    \vspace{-20pt}
  \centering
  \begin{subfigure}{.52\linewidth}
    \includegraphics[width=\linewidth]{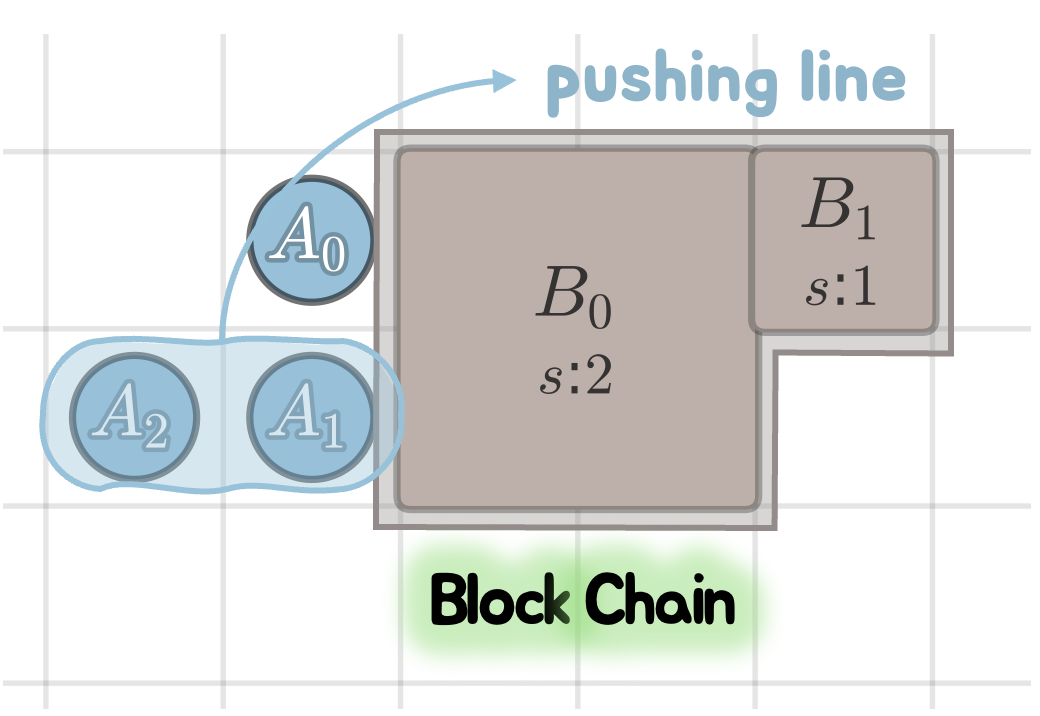}
    \caption{Successful chain}
    \label{fig:succ_chain}
  \end{subfigure}\hfill
  % \vspace{4pt}
  \begin{subfigure}{.47\linewidth}
    \includegraphics[width=\linewidth]{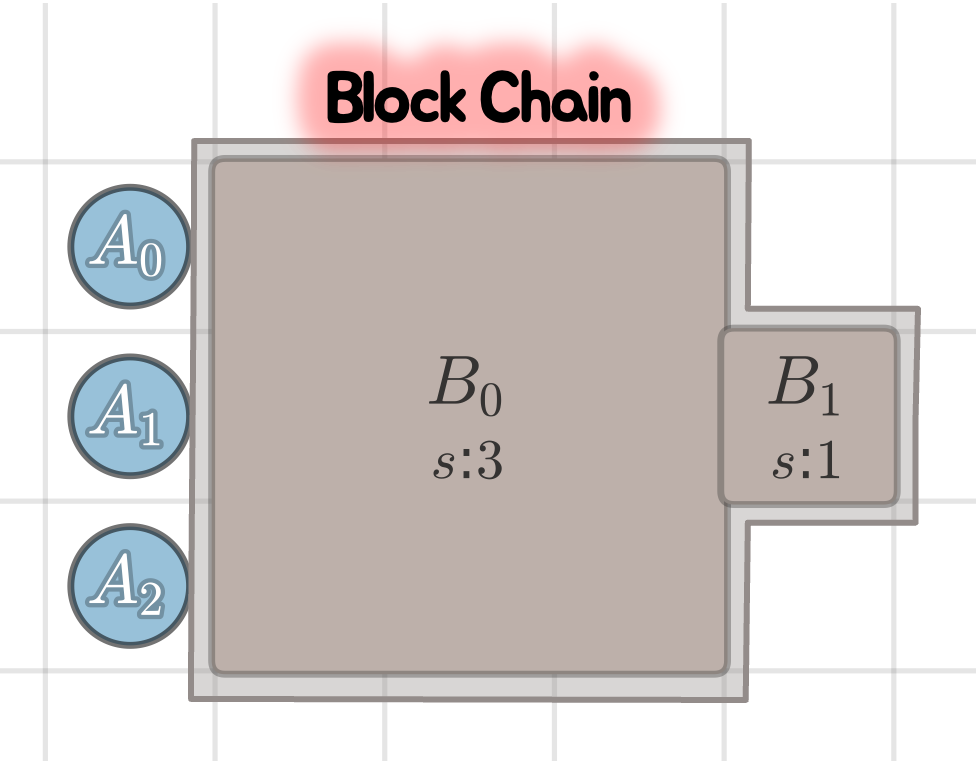}
    \caption{Failed chain}
    \label{fig:fail_chain}
  \end{subfigure}
  \caption{Illustration of chains: a chain succeeds or fails depending on whether available agents meet the block’s weight requirement and the maximum force exerted on the block face.}
  \label{fig:chain_examples}
  \vspace{-15pt}
\end{wrapfigure}

\textbf{Block chain.}
A block $B(g) \in \mathcal{B}$ of weight $w_{b(g)}$ may occupy one or more grid cells. When another block  lies directly in front of $B_j$ along a push direction $d \in \{\uparrow, \downarrow, \leftarrow, \rightarrow\}$, the blocks form a \textit{block chain}. The chain behaves as a composite structure whose motion depends on whether the total applied force at its leading face is sufficient to move all blocks within the chain. Let $i$ index agents in the corresponding \textit{agent chain}, each contributing unit force $f_i = 1$ along direction $d$. A chain advances one cell in direction $d$ if and only if $\sum_i f_i \geq \sum_{B \in \text{chain}} w_{b(g)}$ and all destination cells are unoccupied and within bounds. Upon success, all blocks in the chain advance by one cell; otherwise, the entire chain of blocks remains in place.

\textbf{Agent chain.}
Agents exert unit forces on blocks through pushing chains. A pushing chain forms when multiple agents align \textit{collinearly} behind a block and all push in the direction of the block face. 
%\carlee{``pushing face'' may need to be defined. Instead, we could just say that the all agents push in the direction of the block face} 
The effective force at the contact face is the number of aligned agents in that direction. 
If this total meets or exceeds the required block (or chain) weight, the structure advances. 
Otherwise, the attempt fails and all participating agents remain in place, as shown in Fig.~\ref{fig:chain_examples}.

\textbf{Agent Movement and Collisions}
While the environment supports a dual-layer action interface, primitive and symbolic (see subsection \ref{sec:action_space} for details), each symbolic action is unrolled into a sequence of primitive actions, with all movements and collisions resolved at the primitive level.
At each timestep, each agent issues a primitive action $a_i \in \mathcal{A}$, which specifies a target cell on the grid.
%\carlee{Are there constraints on agent speed? Is this a symbolic action?} 
A move is valid if the target cell lies within bounds and is not occupied by a block 
(including newly moved blocks). If multiple agents attempt to move into the same unoccupied cell, the agent with the smallest index $i$ successfully claims the cell, while all others involved in the conflict remain unmoved. If any agent attempts to move into a cell occupied by a stationary agent, the move fails and both agents stay in place. This rule prevents overlap and introduces a consistent tie-breaking mechanism for simultaneous movements.

\subsubsection{Cognitive Layer Design}

CUBE features a dual-layer interface that integrates symbolic reasoning with vector-based representations across \textbf{observation, action, and feedback channels}. The symbolic layer abstracts the dynamics of the environment into discrete entities and relations such as the distance between the agents and the quorum, allowing high-level reasoning, planning, and language interaction. Complementing this, the vector-based layer provides dense spatial and state features suitable for reinforcement learning and low-level control. Together, these layers allow agents to ground symbolic reasoning in embodied experience, supporting cooperative learning across reinforcement, language, and hybrid agent architectures.

\vspace{5mm}
{\large \textbf{Observation Space}} \label{sec:obs_space}

CUBE provides two observation modalities: a symbolic observation and a multi-channel observation. This dual interface supports diverse agent architectures, enabling reinforcement learning agents to rely on grid-based encodings, LLM-based agents to operate over symbolic state descriptions, or novel approaches that combine both.

\textbf{Symbolic Observation.}
At each step $t$, every agent $i$ receives a symbolic dictionary describing the current state. 
This includes global environment information (grid size, positions of all agents) as well as a compact summary for each block (block ID, weight, position, and distance to the goal column). 
The dictionary also records all symbolic actions taken so far in the episode, along with their corresponding primitive actions and the status (start, in progress, or end) at each timestep. %\carlee{Is a task equivalent to a block?}
This structured interface allows reasoning directly about concepts such as which blocks remain, how far they are from the goal, and where teammates are, supporting high-level planning and coordination.

\textbf{Multi-channel Observation.}
In addition, a five-channel grid encodes agent locations, block weights, the goal column, a channel marking which agent occupies each cell, and a channel marking which block occupies each cell.
This representation resembles standard reinforcement learning observations and is primarily included for compatibility with reinforcement learning pipelines and for visualization.

\vspace{5mm}
{\large \textbf{Action Space} }
\label{sec:action_space}

CUBE supports two sets of action spaces. 
The primitive action space provides low-level grid movements, enabling agents to interact directly with the environment through discrete directional moves. 
The symbolic action space abstracts these primitives into higher-level cooperative strategies, such as aligning on a block face, synchronizing for a push, or waiting for teammates. 
Together, these two levels allow experiments to target both reinforcement learning agents, which operate naturally over primitive actions, and LLM-based agents, which benefit from reasoning over symbolic actions.

\textbf{Primitive Actions.}
Each agent selects from a discrete 5-action set
\[
\mathcal{A}=\{\texttt{STAY}=0,\ \texttt{UP}=1,\ \texttt{DOWN}=2,\ \texttt{LEFT}=3,\ \texttt{RIGHT}=4\}.
\]
At time $t$, each agent $i$ issues an action $a_{i,t}$ specifying a movement direction, and all agents act in parallel to move one unit in their respective directions.
Moves succeed only if the target cell is free; collisions with walls, agents, or insufficiently supported blocks cause the agent to remain in place. A push succeeds if the aligned agents' combined force exceeds the total weight of the aligned blocks and the destination cell is free, in which case both the blocks and agents advance one step.

\textbf{Symbolic Actions.}
Beyond primitive grid movements, CUBE provides a small library of \textit{symbolic actions} that capture essential coordination behaviors while remaining easy to ground in the environment. Each symbolic action is executed by compiling it into a sequence of primitive moves until the specified condition is met. In our final setup, we use only three symbolic actions: \texttt{move}, \texttt{wait}, and \texttt{push}. \texttt{move} compiles into repeated primitive movements in a fixed direction, \texttt{wait} compiles into \texttt{STAY} actions for a fixed number of steps, and \texttt{push} compiles into repeated primitive pushes against the specified block until the push terminates.

\begin{table}[h]
\scriptsize
\renewcommand{\arraystretch}{1.2}
\setlength{\tabcolsep}{6pt}
\centering
\begin{tabularx}{\linewidth}{@{}l m{0.22\linewidth} m{0.60\linewidth}@{}}
\toprule
\textbf{Action} & \textbf{Arguments} & \textbf{Effect} \\
\midrule
%\texttt{noop} & (steps) & Do nothing for one step (single-step action). \\
\texttt{move} & (\textit{direction}, \textit{steps}) & Move in the specified direction for a fixed number of steps (compiled into primitive moves). \\
\texttt{wait} & (\textit{steps}) & Remain idle for a fixed number of steps (compiled into \texttt{STAY}). \\
\texttt{push} & (\textit{block\_id}, \textit{steps}) & Push against the specified block by repeatedly issuing primitive actions directed into the block face; succeeds only if enough agents align and apply sufficient force. \\
\bottomrule
\end{tabularx}
\caption{Symbolic actions in CUBE used in our experiments. Each action specifies a high-level effect that is decomposed into primitive actions.}
\label{tab:symb_actions}
\end{table}

Although the symbolic action set is compact, these actions remain expressive because coordination emerges through agents’ choices of where to move, when to wait, and which block to push, all under embodied constraints from collisions, alignment, and quorum requirements.

\subsubsection{Environment Prompt}

    Similarly to the MA-Crafter, the environment prompt outlines the basic rules, actions, and constraints of the \texttt{CUBE} world. The prompt describes cooperative block-pushing mechanics, where blocks of different weights require different numbers of agents to push simultaneously.\\
    
    \begin{tcolorbox}[breakable, colback=gray!2, colframe=black!60, title=\texttt{CUBE} Environment Specification, fonttitle=\bfseries]
        \label{prompt:env}
        \textbf{Goal} \\
        Push all blocks to the goal column (rightmost column) to deliver them. Blocks require multiple agents pushing simultaneously if block weight $> 1$. \\
        
        \textbf{Grid} \\
        \hspace*{1em}-- Simple $K \times K$ grid (default $8 \times 8$) \\
        \hspace*{1em}-- Goal column is the rightmost column (column $K-1$) \\
        \hspace*{1em}-- Agents start at various positions on the left side \\
        \hspace*{1em}-- Blocks are placed in the middle area \\
        
        \textbf{Blocks} \\
        Each block has: \\
        \hspace*{1em}-- \textbf{ID}: Unique identifier \\
        \hspace*{1em}-- \textbf{Weight}: Determines size (weight $\times$ weight square) and push requirement \\
        \hspace*{1em}-- \textbf{Position}: (row, col) of top-left corner \\
        
        \textbf{Push Requirements} \\
        \hspace*{1em}-- Weight 1: 1 agent can push alone \\
        \hspace*{1em}-- Weight 2: 2 agents must push simultaneously from same side \\
        \hspace*{1em}-- Weight 3: 3 agents must push simultaneously from same side \\
        \hspace*{1em}-- etc. \\
        
        \textbf{Cooperative Mechanics} \\
        To successfully push a block of weight $W$: \\
        \hspace*{1em}1. $W$ agents must be adjacent to the \textbf{SAME side} of the block \\
        \hspace*{1em}2. All $W$ agents must issue push/move commands \textbf{INTO} the block simultaneously \\
        \hspace*{1em}3. If successful, both the block and pushing agents move forward \\
        \hspace*{1em}4. Block is delivered when any part reaches the goal column \\
        
        \textbf{Actions} \\
        \texttt{move(direction, num\_steps)}: Move in a direction for \texttt{num\_steps}. \\
        \hspace*{1em}-- \texttt{direction}: ``up'', ``down'', ``left'', ``right'' \\
        \hspace*{1em}-- \texttt{num\_steps}: 1--10 \\
        
        \texttt{push(block\_id, num\_steps)}: Push a block by moving into it. \\
        \hspace*{1em}-- \texttt{block\_id}: ID of the target block \\
        \hspace*{1em}-- \texttt{num\_steps}: How many steps to push \\
        \hspace*{1em}-- Direction is automatic based on agent position relative to block \\
        \hspace*{1em}-- Agent must be adjacent to the block \\
        
        \texttt{wait(num\_steps)}: Do nothing for \texttt{num\_steps}. \\
        \hspace*{1em}-- Useful for synchronizing with other agents \\
        \hspace*{1em}-- \texttt{num\_steps}: 1--10 \\
        
        \textbf{Observation Channels} \\
        5-channel observation ($K \times K \times 5$): \\
        \hspace*{1em}-- Channel 0: Agent mask (1.0 at agent positions) \\
        \hspace*{1em}-- Channel 1: Block weights (value = weight at block cells) \\
        \hspace*{1em}-- Channel 2: Goal strip (1.0 in rightmost column) \\
        \hspace*{1em}-- Channel 3: Agent IDs (+1) \\
        \hspace*{1em}-- Channel 4: Block IDs (+1) \\
        
        \textbf{Rewards} \\
        \hspace*{1em}-- Step cost: Small negative reward per step \\
        \hspace*{1em}-- Delivery reward: Positive reward when block reaches goal \\
        \hspace*{1em}-- Episode ends when all blocks delivered or max steps reached \\
        
        \textbf{Strategy Tips} \\
        1. \textbf{Communication}: Announce which block you're targeting and from which side \\
        2. \textbf{Coordination}: For weight-2+ blocks, agree on timing (e.g., ``push at step 10'') \\
        3. \textbf{Positioning}: Get into position on the same side before pushing \\
        4. \textbf{Use wait}: Synchronize pushes with wait actions \\
        5. \textbf{Efficiency}: Push lighter blocks first if they're blocking heavier ones \\
        \end{tcolorbox}
     %cluster 3: envs
% %\input{appendix/maeil}

% %\input{appendix/metrics}

% % \input{appendix/scalability}

\newpage

\section{Additional Results and Qualitative Analysis}
This cluster provides additional COOP$^2$ traces beyond those shown in the main paper and qualitative analyses of the communication patterns that emerge across different settings.

\subsection{COOP$^2$ Traces}\label{app:extra-trace}

We provide additional process-level traces beyond the 3-agent MA-Crafter results shown in the main part of the paper. The traces below cover the 6-agent setting and CUBE runs, allowing more detailed inspection of how cooperation unfolds across team sizes and environments.

\begin{figure}[H]
    \centering
    \includegraphics[width=0.9\linewidth]{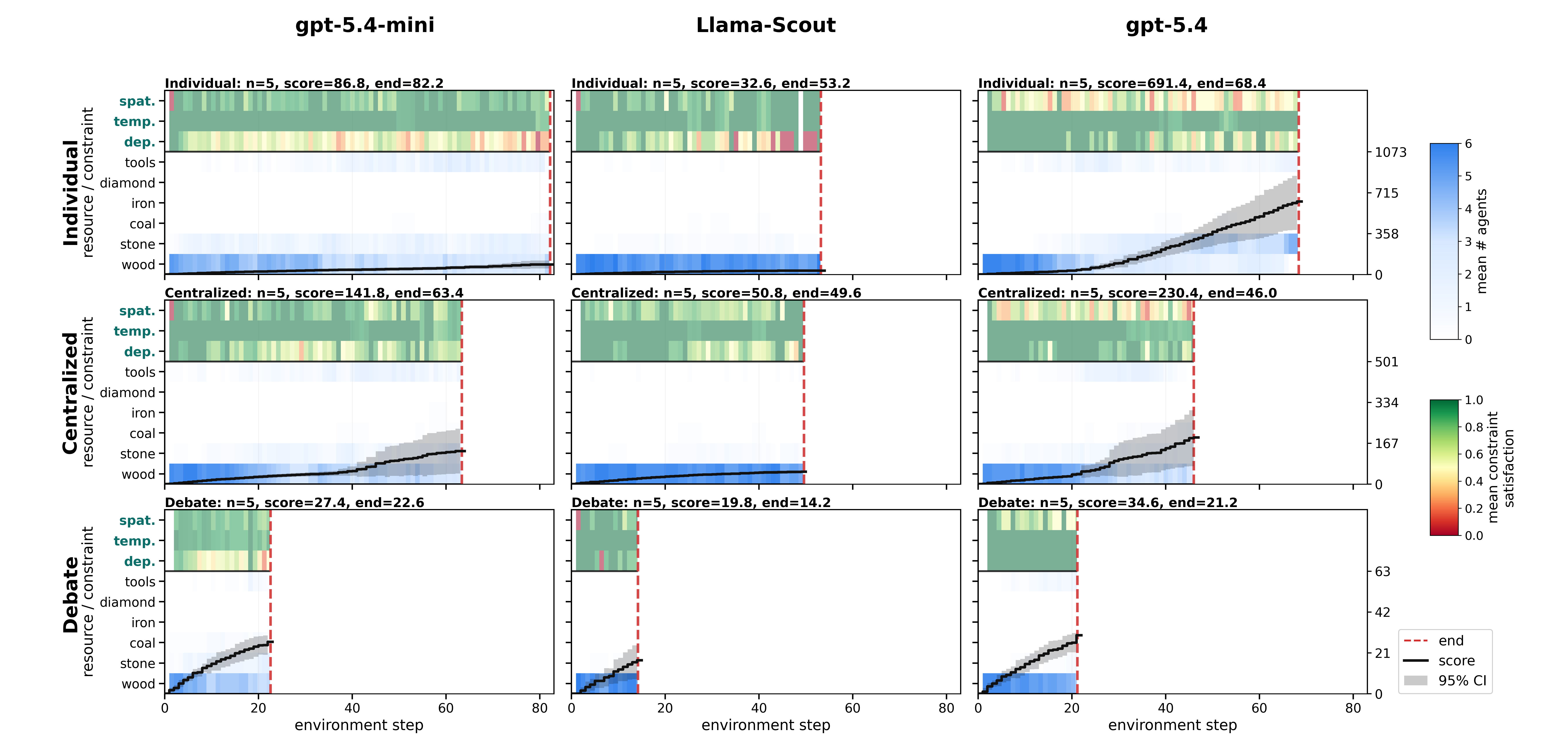}
    \caption{COOP$^2$ process traces for MA-Crafter with 6-agent teams, averaged over five runs. The traces show active task focus, observed constraint satisfaction at collection attempts, and cumulative team score across backbones and communication structures.}
    \label{fig:placeholder}
\end{figure}

\begin{figure}[H]
    \centering
    \includegraphics[width=0.9\linewidth]{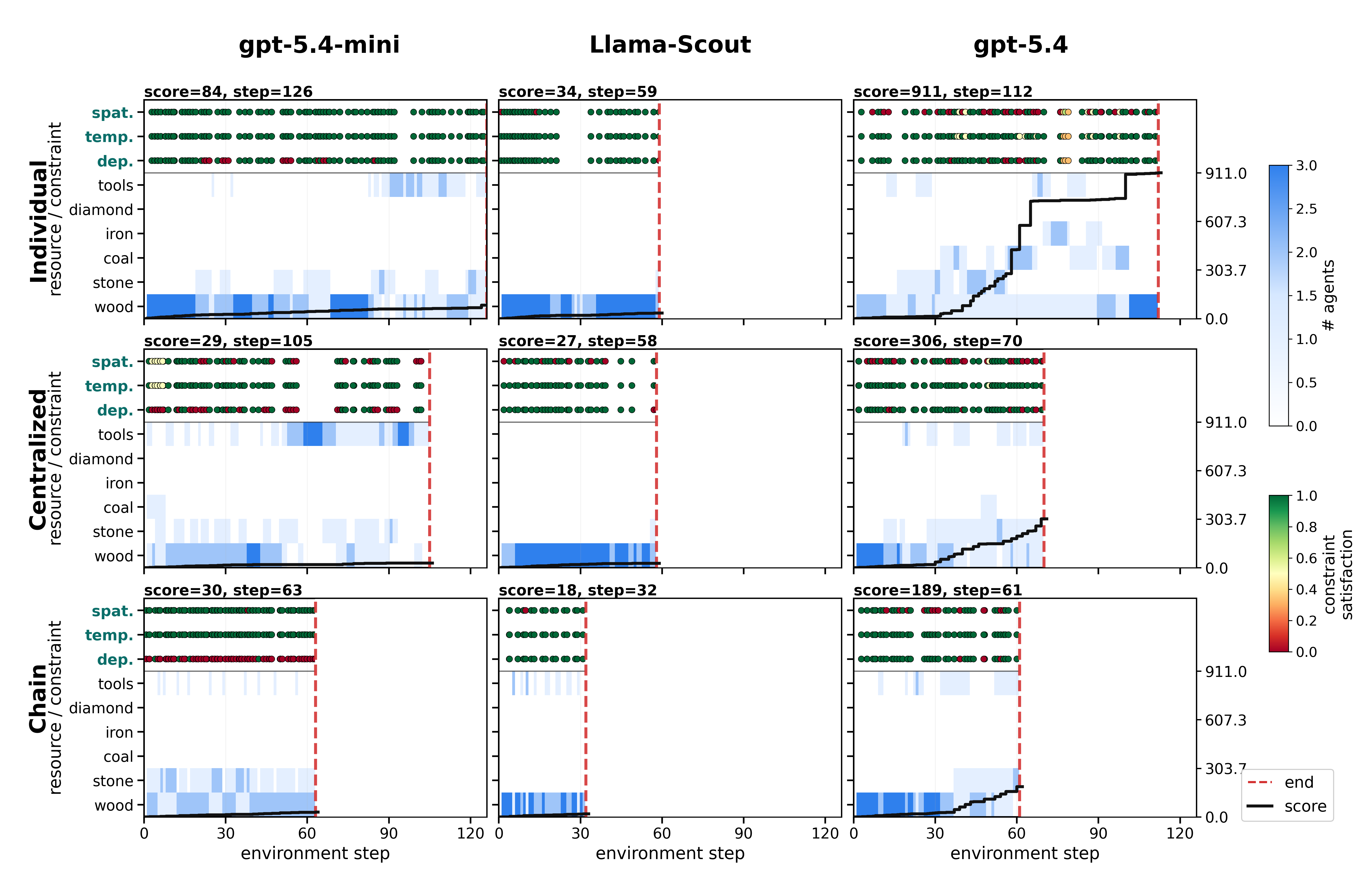}
    \caption{COOP$^2$ process traces for MA-Crafter with 3-agent teams, single run. The traces show how agents' active task focus, constraint satisfaction, and cumulative score evolve over time, providing an additional view of cooperation dynamics in a spatial coordination environment.}
    \label{fig:placeholder}
\end{figure}

\begin{table*}[t]
\centering
\scriptsize
\setlength{\tabcolsep}{4pt}
\begin{tabular}{lccccccccccc}
\toprule
Structure
& \begin{tabular}{c} Score \\ $\uparrow$ \end{tabular}
& Steps
& Score/Step
& Plans/Agent
& Msg.
& Intr.
& Total Dec.
& Spat. Viol.
& Temp. Viol.
& Dep. Viol. \\
\midrule

\multicolumn{11}{l}{\textbf{3-agent GPT-5.4-mini}} \\
Individual  & 3.0 & 116.0 & 0.03 & 30.7 & 0.0  & 0.0  & 137.0 & 0.20 & 0.26 & 0.00 \\
Centralized & 3.0 & 80.0  & 0.04 & 23.3 & 54.0 & 29.0 & 183.9 & 0.27 & 0.30 & 0.00 \\
Chain       & 1.0 & 54.0  & 0.02 & 23.7 & 48.0 & 46.0 & 197.8 & 0.17 & 0.23 & 0.00 \\

\midrule
\multicolumn{11}{l}{\textbf{3-agent GPT-5.4}} \\
Individual  & 1.0 & 78.0 & 0.01 & 16.3 & 0.0  & 0.0  & 130.1 & 0.29 & 0.48 & 0.00 \\
Centralized & 1.0 & 52.0 & 0.02 & 12.3 & 39.0 & 24.0 & 215.3 & 0.23 & 0.34 & 0.00 \\
Chain       & 1.0 & 37.0 & 0.03 & 12.3 & 24.0 & 19.0 & 199.5 & 0.46 & 0.46 & 0.00 \\

\midrule
\multicolumn{11}{l}{\textbf{6-agent GPT-5.4-mini}} \\
Individual  & 3.0 & 86.0 & 0.03 & 21.7 & 0.0   & 0.0  & 192.3 & 0.10 & 0.12 & 0.00 \\
Centralized & 0.0 & 75.0 & 0.00 & 20.5 & 126.0 & 97.0 & 342.3 & 0.10 & 0.64 & 0.00 \\
Chain       & 3.0 & 39.0 & 0.08 & 13.5 & 66.0  & 52.0 & 310.8 & 0.14 & 0.22 & 0.00 \\

\midrule
\multicolumn{11}{l}{\textbf{6-agent GPT-5.4}} \\
Individual  & 1.0 & 49.0 & 0.02 & 15.0 & 0.0  & 0.0  & 233.5 & 0.21 & 0.41 & 0.00 \\
Centralized & 6.0 & 59.0 & 0.10 & 10.7 & 55.0 & 40.0 & 324.3 & 0.41 & 0.55 & 0.00 \\
Chain       & 1.0 & 23.0 & 0.04 & 6.7  & 35.0 & 28.0 & 332.9 & 0.31 & 0.33 & 0.00 \\

\bottomrule
\end{tabular}
\caption{CUBE cooperation results.}
\label{tab:cube-coop2-repair-off-seed42}
\end{table*}

\begin{figure}[H]
    \centering
    \includegraphics[width=1\linewidth]{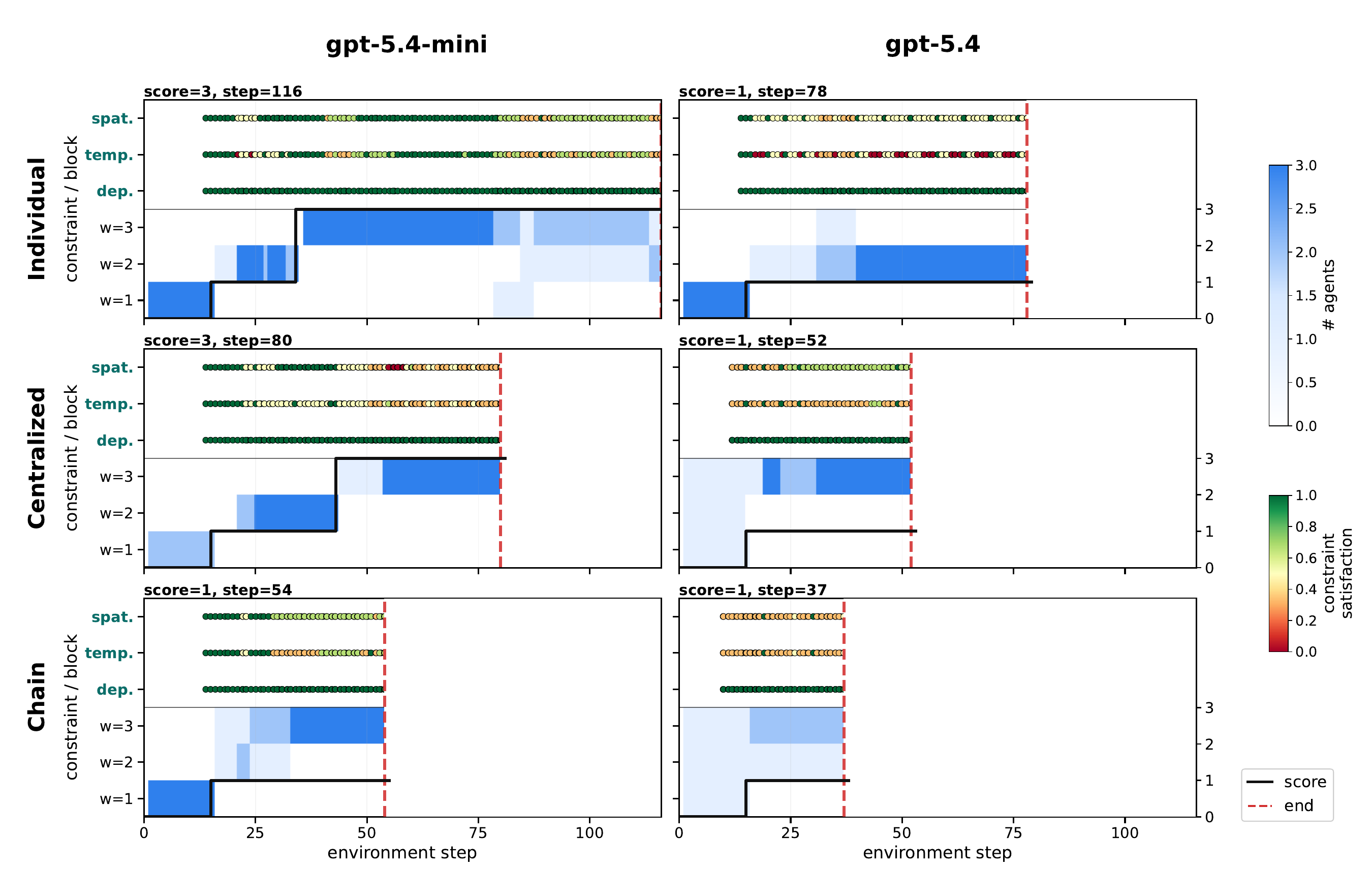}
    \caption{COOP$^2$ process traces for CUBE with 3-agent teams, single run. The traces show how agents' active task focus, constraint satisfaction, and cumulative score evolve over time, providing an additional view of cooperation dynamics in a spatial coordination environment.}
    \label{fig:placeholder}
\end{figure}

\begin{figure}[H]
    \centering
    \includegraphics[width=1\linewidth]{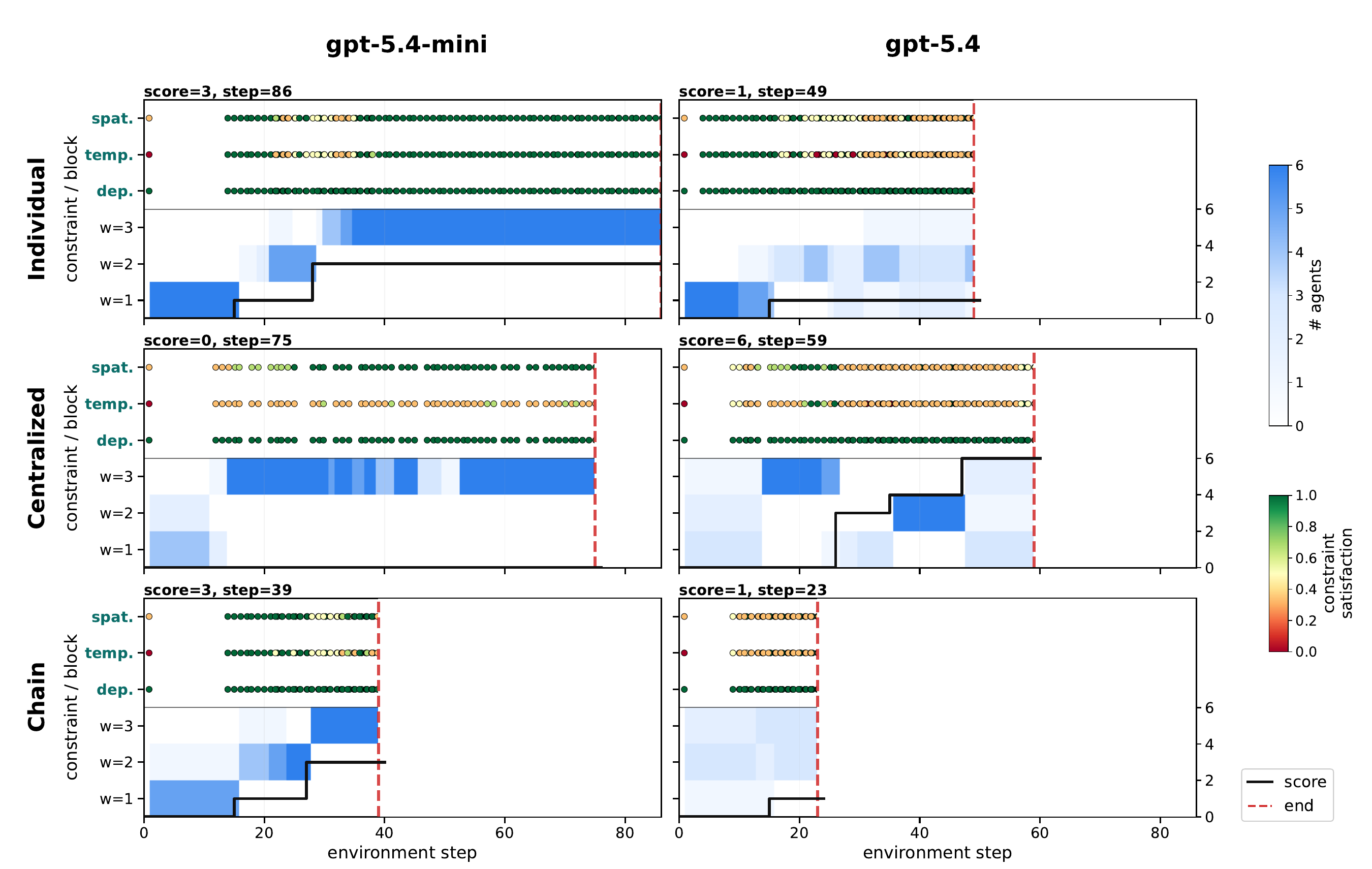}
    \caption{COOP$^2$ process traces for CUBE with 6-agent teams, single run. The traces show how agents' active task focus, constraint satisfaction, and cumulative score evolve over time, providing an additional view of cooperation dynamics in a spatial coordination environment.}
    \label{fig:placeholder}
\end{figure}

\subsection{Annotated Communications}
%\graphicspath{{rebuttal/}}
We provide annotated examples of agent communication under different structures and difficulty levels, illustrating common patterns in how LLM agents recognize, discuss, and attempt to satisfy cooperative constraints. These examples qualitatively complement the aggregate metrics and traces in the main paper.

\begin{figure}[H]
    \centering
    \includegraphics[width=0.9\linewidth]{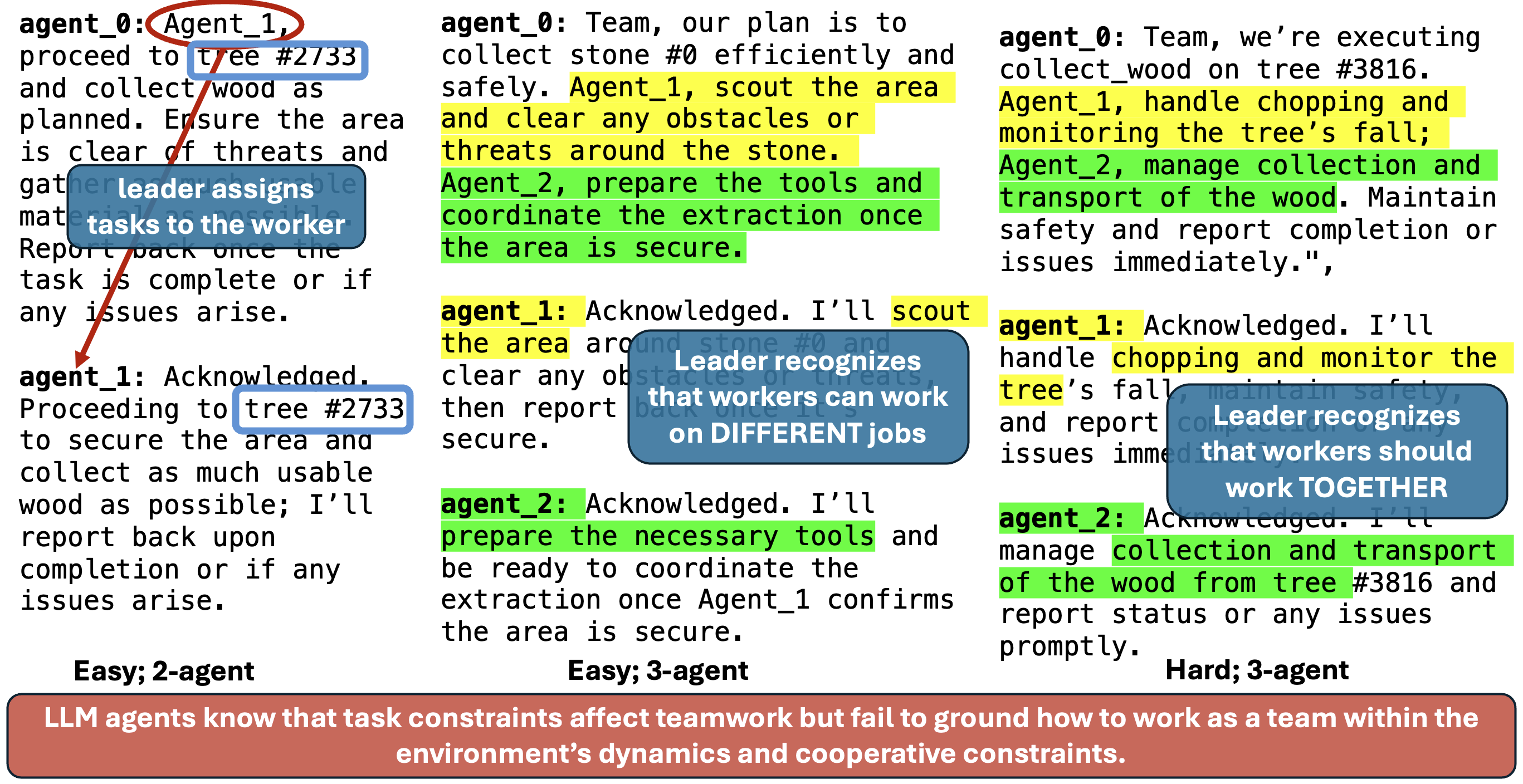}
    \caption{Communication messages of LLM-MAS operating under centralized communication structures. The left panel shows a 2-agent setting in Easy mode, the middle panel shows 3 agents in Easy mode, and the right panel shows 3 agents in Hard mode. In the Easy setting, the leader assigns distinct tasks to workers. In Hard settings, although agents recognize the presence of constraints, they struggle to determine how to satisfy them. The leader is aware that coordination is required, but lacks grounding in how to execute effective teamwork within the environment.}
\end{figure}

\begin{figure}[H]
    \centering
    \includegraphics[width=0.8\linewidth]{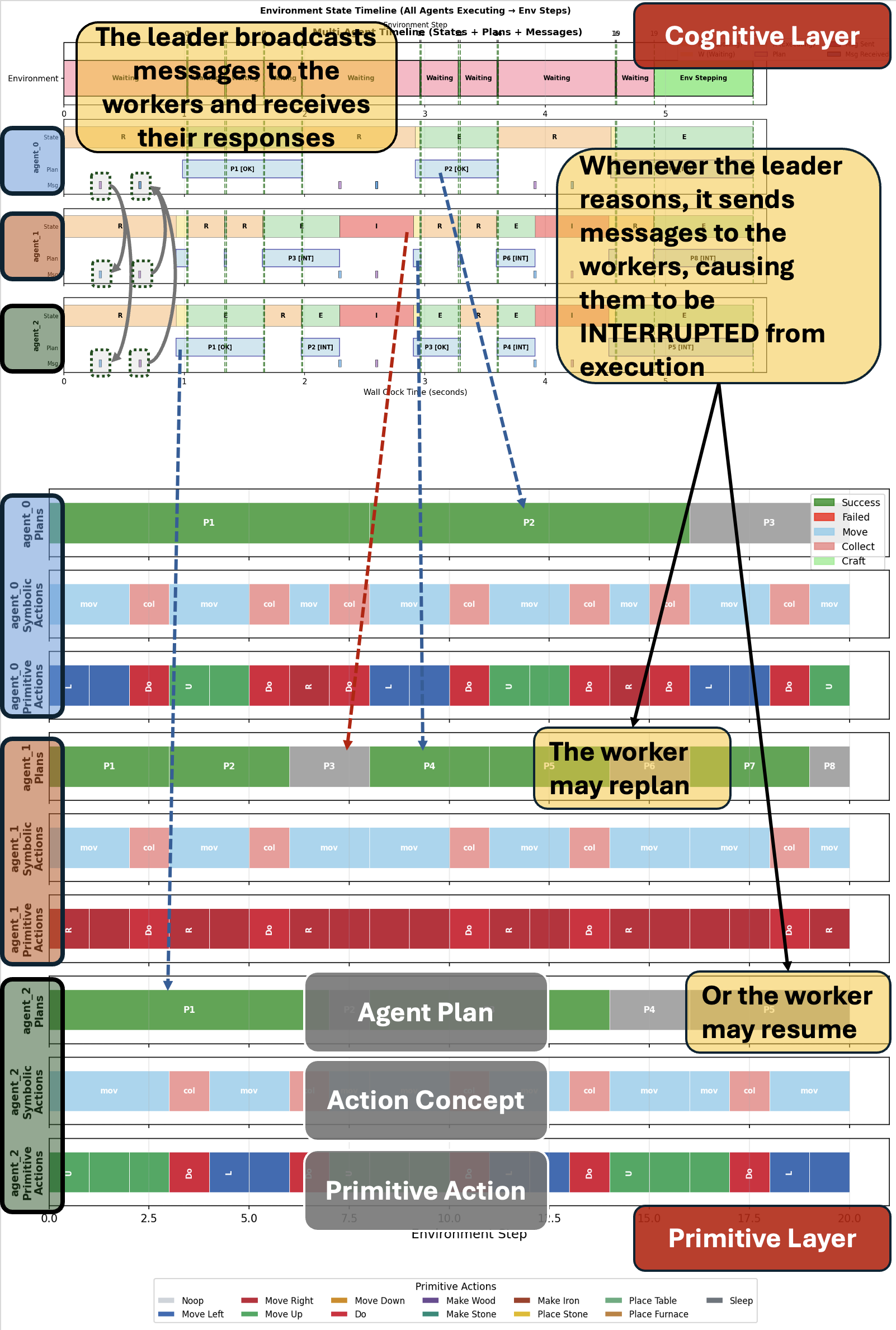}
    \caption{COOP$^2$ bridges the cognitive layer, where agents perform reasoning, planning, and communication, and the primitive layer, where plans are grounded into primitive actions. The figure depicts a centralized communication structure under COOP$^2$: the leader broadcasts messages to workers, interrupting workers’ ongoing execution and triggering replanning or resumption.}
\end{figure}

% \begin{figure}[h]
%     \centering
%     \includegraphics[width=0.9\linewidth]{topo-messages.png}
%     \caption{Communication messages of LLM-MAS operating under decentralized and debate communication structures. Decentralized settings (left) lead to unfocused discussions, making it difficult for agents to make progress. Debate settings (right) lead to more diverse plans, which may surface potential issues or stall teamwork.}
%     \label{fig:placeholder}
% \end{figure}

\begin{figure}[H]
    \centering
    \includegraphics[width=0.75\linewidth]{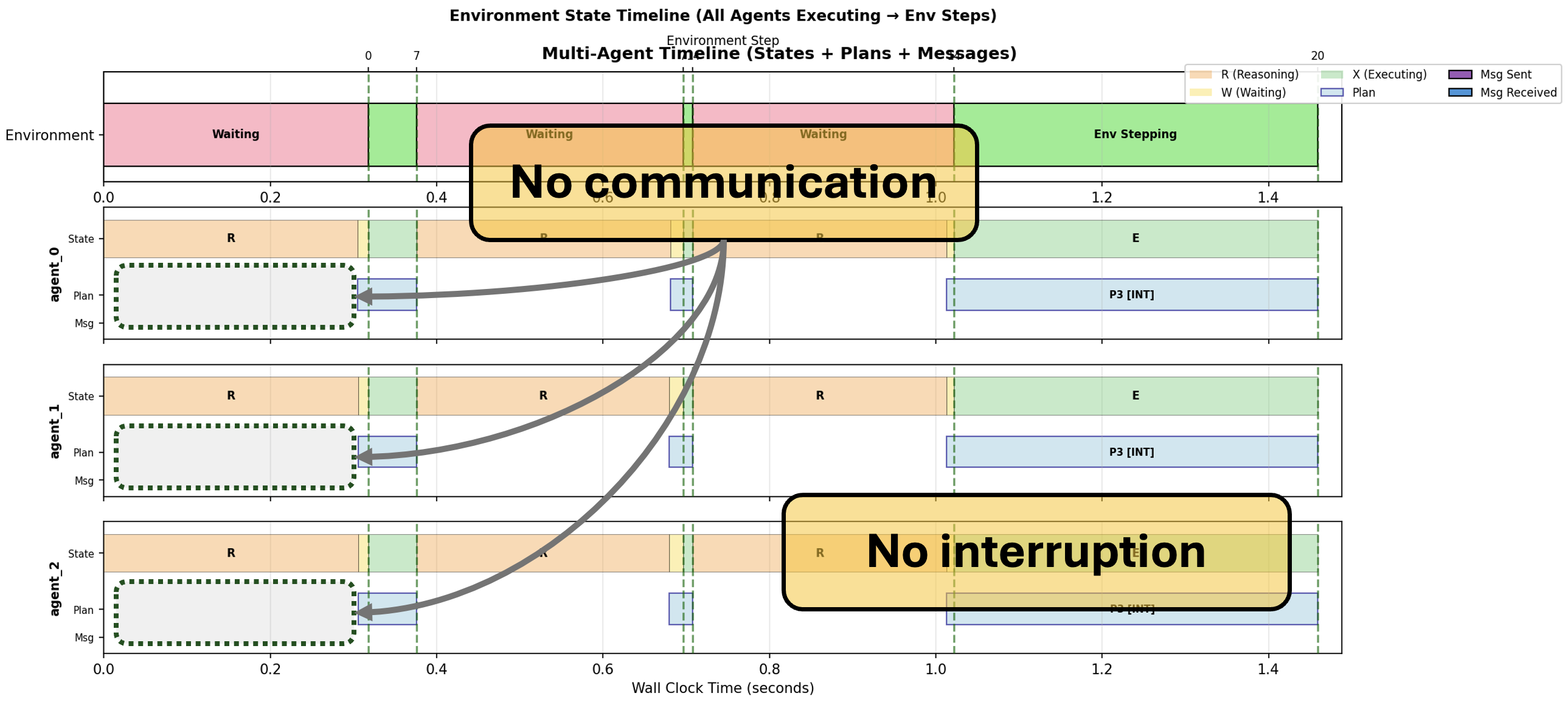}
    \caption{Individual communication structure.}
    \label{fig:placeholder}
\end{figure}

\begin{figure}[H]
    \centering
    \includegraphics[width=0.75\linewidth]{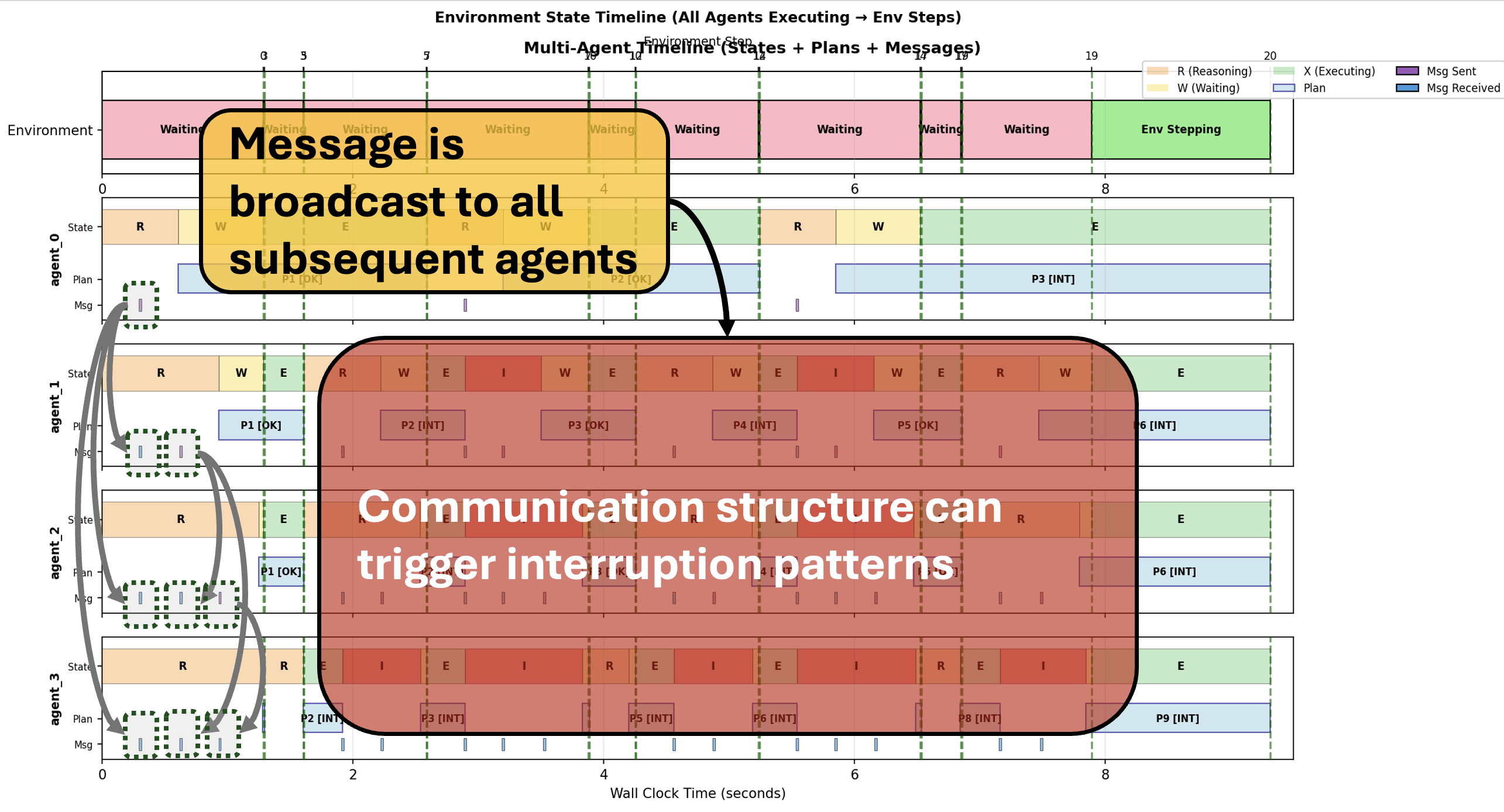}
    \caption{Chain communication structure.}
    \label{fig:placeholder}
\end{figure} %cluster 4: add rsult, trace, annotation

% \input{appendix/comm_structures}

% \input{appendix/response_template}

% \input{appendix/prompts}

% \input{appendix/annotated-comm}

%% \input{appendix/cube}

%\input{appendix/more_results}

% \section{Technical appendices and supplementary material}
% Technical appendices with additional results, figures, graphs, and proofs may be submitted with the paper submission before the full submission deadline (see above). You can upload a ZIP file for videos or code, but do not upload a separate PDF file for the appendix. There is no page limit for the technical appendices. 

% Note: Think of the appendix as ``optional reading'' for reviewers. The paper must be able to stand alone without the appendix; for example, adding critical experiments that support the main claims to an appendix is inappropriate. 

% %%%%%%%%%%%%%%%%%%%%%%%%%%%%%%%%%%%%%%%%%%%%%%%%%%%%%%%%%%%%

% \newpage
%\input{checklist.tex}

\end{document}